\documentclass{article}

\usepackage[main, final, nonatbib]{neurips_2026}

\usepackage[utf8]{inputenc} 
\usepackage[T1]{fontenc}    
\usepackage{hyperref}       
\usepackage{url}            
\usepackage{booktabs}       
\usepackage{amsfonts}       
\usepackage{nicefrac}       
\usepackage{microtype}      
\usepackage{xcolor}         

\usepackage[numbers]{natbib}

\usepackage{amsmath}
\usepackage{graphicx}
\usepackage{multirow}

\usepackage{pifont}

\usepackage[capitalize]{cleveref}

\title{Beyond the Grid: Continuous Dictionary Pursuit for Interpretable Signal Decomposition}

\author{%
  Nick~Janssen\\
  Institute of Information Processing\\
  Leibniz University Hannover\\
  \texttt{janssen@tnt.uni-hannover.de} \\
  \And
  Melanie~Schaller\\
  Institute of Information Processing\\
  Leibniz University Hannover\\
  \texttt{schaller@tnt.uni-hannover.de} \\
  \And
  Bodo~Rosenhahn\\
  Institute of Information Processing\\
  Leibniz University Hannover\\
  \texttt{rosenhahn@tnt.uni-hannover.de} \\
}

\begin{document}

\maketitle

\begin{abstract}
Deep learning models achieve high accuracy in signal processing but remain difficult to interpret. 
Classical matching pursuit and sparse coding offer better transparency but suffer from spectral leakage and grid mismatch when signal frequencies do not align with discrete dictionary atoms. 
We introduce Continuous Dictionary Pursuit (CDP), a framework for interpretable signal decomposition that learns the parameters of closed-form functions directly from data.
CDP optimizes over a continuous manifold of differentiable atoms and, unlike prior continuous pursuit methods that operate on a single analytic family, searches a heterogeneous library spanning periodic, trend, transient, and discontinuous structures.
The algorithm employs a greedy iterative strategy using data-driven priors and gradient-based optimization to isolate individual signal components, with discontinuous waveforms made differentiable through spectral sharpness annealing.
To ensure a sparse atom set for reconstruction, the framework utilizes a stopping criterion based on a predefined atom budget and residual error convergence.
Evaluations on synthetic and real-world datasets show that CDP removes the grid discretization error for atom families contained in its library, recovers generative parameters to within $1\%$ on stationary synthetic signals expressible in that library, and achieves the best sparsity-accuracy trade-off among classical, continuous-pursuit, and neural baselines in the sparse regime.
Beyond decomposition, we demonstrate that the extracted atoms can serve as structured curriculum targets for training sequence models, improving forecasting accuracy over standard end-to-end training on most evaluated datasets.
The method offers a glass-box alternative for signal decomposition by providing high reconstruction fidelity alongside direct physical interpretability.
\end{abstract}


\section{Introduction}
\label{sec:introduction}
The decomposition of univariate time series into interpretable components is a fundamental problem across a wide range of scientific and engineering disciplines, from energy consumption planning~\cite{forootani2024_climateawaredeepneural,mustafa2025_energyforecasting,zhou2022_fedformer} and financial market analysis~\cite{chakraborti2007_financialoverview,tsay2005_financeanalysis} to medical signal processing~\cite{bui2018_cardiocasculardiseases,rios1996_typhoidfever,sreenivas2021_arrhythmiacnn} and industrial sensor networks~\cite{farahani2025_manufacturingsystems,schaller2026_s4dbio}. 
Across these domains, the ability to express a signal as a sum of physically meaningful components is often as valuable as predictive accuracy, since the decomposition itself enables diagnosis, anomaly attribution, and scientific understanding.

With increasing signal complexity, deep learning approaches achieve strong results on reconstruction benchmarks yet function as black boxes, offering little insight into the underlying signal structure~\cite{rudin2019_blackbox}.
Interpretable-by-design models address this by construction, for example through component-based image classifiers~\cite{norrenbrock2025_chiqpm}, and CDP pursues the same principle for signal decomposition.
This matters in safety-critical settings such as healthcare~\cite{da2021_clinicalinterpretability,singh2020_medicalimageexplainable} and autonomous systems~\cite{jing2022_inaction,ozkurt2024_interpretablechaos,yan2023_selfinterpretable}, where understanding a model's output is a practical and increasingly regulatory requirement~\cite{doshi2017_towardsinterpretable,goodman2017_euregulations}.

Classical approaches face a different set of limitations. 
Methods such as STL~\cite{cleveland1990_stl} produce components with clear semantic labels (trend, seasonality, residual) but no explicit functional form, while sparse coding methods such as Orthogonal Matching Pursuit (OMP)~\cite{pati1993_omp} represent signals as linear combinations of dictionary atoms and offer explicit, term-by-term reconstructions. 
However, operating on a fixed discrete dictionary exposes them to the grid mismatch problem~\cite{duarte2013_spectralcompressivesensing,pezeshki2015_gridmismatch}: when a signal component's true parameters fall between grid points, the reconstruction requires multiple atoms to approximate a single physical feature, inflating model complexity and destroying interpretability.

\begin{figure}[htbp]
  \begin{center}
    \includegraphics[width=0.9\columnwidth]{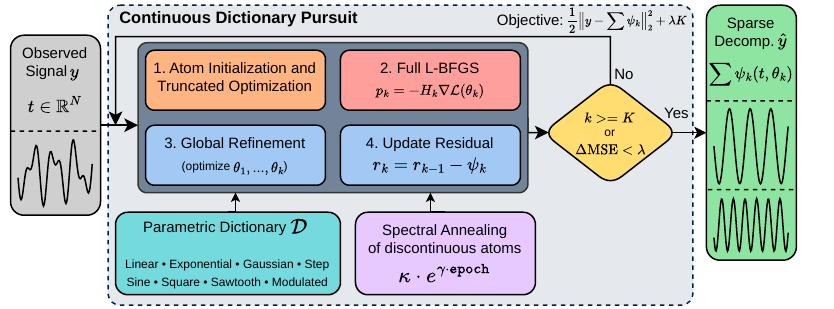}
    \caption{\textbf{Continuous Dictionary Pursuit:} The iterative greedy procedure initializes atoms from data-driven priors, identifies a candidate via truncated optimization, fully optimizes it with L-BFGS~\cite{liu1989_lbfgs}, and adds it to the additive model until the atom budget is exhausted or the residual error stagnates.}
    \label{fig:cdp_diagram}
  \end{center}
\end{figure}%
We propose Continuous Dictionary Pursuit (CDP), illustrated in \autoref{fig:cdp_diagram}, a greedy iterative framework that optimizes directly over a continuous manifold of parameterized closed-form functions, removing the grid discretization error for atom families contained in its library through FFT-based spectral priors, truncated parallel optimization, and a global joint refinement step.
We evaluate CDP against nine baselines ranging from classical decomposition to deep learning, including the closest continuous pursuit methods NOMP and SFW~\cite{mamandipoor2016_nomp,denoyelle2019_sfw}, demonstrating its utility as a glass-box model in which every component has a precise mathematical definition and a direct physical interpretation.

The contributions of this work are as follows:
\begin{itemize}
    \item \textbf{Continuous Dictionary Pursuit (CDP):} A greedy decomposition framework that optimizes over a continuously parameterized manifold of closed-form functions spanning periodic, trend, transient, and discontinuous waveform families, removing the grid discretization error of fixed-dictionary methods for atom families contained in its library.
    \item \textbf{Greedy Forward Selection:} A hybrid selection strategy combining FFT-based spectral priors for candidate initialization with truncated parallel optimization and full L-BFGS refinement under soft barrier penalties.
    \item \textbf{Pareto-Efficient Reconstruction:} On real-world LTSF benchmarks, CDP achieves the best sparsity-accuracy trade-off among all evaluated baselines in the sparse regime.
    \item \textbf{Atom-Based Curriculum Learning:} CDP atoms serve as structured curriculum targets for sequence models, reducing forecasting error over standard end-to-end training.
    \item \textbf{Open Source Release:} The full implementation of CDP is publicly available on GitHub.\footnote{\url{https://github.com/nick-janssen/ContinuousDictionaryPursuit}}
\end{itemize}


\section{Related Work}
\label{sec:related_works}
\textbf{Classical and Spectral Decomposition:}
\label{sec:classical_statistical_decomposition}
The ARIMA family~\cite{box1976_arima} and exponential smoothing produce closed-form representations with statistically interpretable coefficients, but their linearity limits their ability to capture non-stationary dynamics~\cite{zhang2003_arimaforecasting}.
STL~\cite{cleveland1990_stl} and MSTL~\cite{bandara2025_mstl} separate signals into seasonal, trend, and remainder components via iterative LoESS smoothing, yielding semantically labeled but non-closed-form expressions, and Prophet~\cite{taylor2018_prophet} fits piecewise linear trends with Fourier seasonalities yet cannot adapt to non-sinusoidal or aperiodic behavior.
Subspace methods such as SSA~\cite{vautard1992_ssa} and DMD~\cite{schmid2010_dmd} decompose the trajectory matrix via Singular Value Decomposition, but their components are linear combinations of lagged observations that require additional analysis to link to physical phenomena.
CDP instead learns an additive functional form whose components are members of named parametric families with directly readable parameters.

\textbf{Sparse Coding and Dictionary Learning:}
\label{sec:sparse_coding_dict_learning}
Matching Pursuit (MP)~\cite{mallat1993_matchingpursuit} and Orthogonal Matching Pursuit (OMP)~\cite{pati1993_omp} greedily represent signals as sparse linear combinations of dictionary atoms, and K-SVD~\cite{aharon2006_ksvd} further adapts the atoms to data, though they remain unstructured and physically uninterpretable.
All fixed-dictionary methods share the grid mismatch problem~\cite{duarte2013_spectralcompressivesensing,pezeshki2015_gridmismatch}: when a component's true frequency falls between grid points, several neighboring atoms are required to approximate a single physical feature, inflating model complexity and degrading interpretability.
Atomic norm minimization~\cite{bhaskar2013_atomicnorm} addresses this for the sinusoidal case with exact recovery guarantees~\cite{tang2013_compressed,candes2014_superres}, but the underlying Toeplitz structure has no counterpart for mixed atom families containing Gaussians, polynomials, or non-sinusoidal waveforms.

\textbf{Structured, Analytic, and Continuous Dictionaries:}
\label{sec:structured_dictionaries}
A rich body of work imposes structure on learned dictionaries~\cite{dumitrescu2018_dictionarylearning}, learns analysis-model operators instead of synthesis atoms~\cite{rubinstein2013_analysisksvd}, adapts dictionaries across domains~\cite{qiu2012_domainadaptive}, or constrains representations geometrically~\cite{tasissa2023_kdeepsimplex}.
Morphological component analysis (MCA)~\cite{starck2005_mca,bobin2007_mca} separates signals over unions of analytic dictionaries, such as Fourier and wavelet bases, exploiting the morphological distinctness of the sub-dictionaries.
In all of these approaches the learned or selected atom is ultimately a numerical vector, either learned under structural constraints or drawn from a parametric family discretized onto a grid, so interpretation requires reading structure out of the vector after the fact.
Closest to our work are continuous pursuit methods that refine atom parameters off the grid: Newtonized OMP~\cite{mamandipoor2016_nomp} augments greedy frequency estimation with Newton refinement over the continuum, and the Sliding Frank-Wolfe algorithm~\cite{denoyelle2019_sfw} solves the continuous sparse approximation problem (BLASSO) with exact local parameter updates.
Both operate on a single smooth analytic family, which is what makes these exact update steps available.
CDP belongs to this class of continuous pursuit methods and extends it to a heterogeneous library of periodic, trend, transient, and discontinuous waveforms with the atom type selected per iteration (\autoref{sec:methodology}), and both methods serve as baselines in our evaluation (\autoref{sec:reconstruction_eval}).

\textbf{Deep Sequence Models:}
\label{sec:deep_sequence_models}
N-BEATS~\cite{oreshkin2019_nbeats} decomposes signals into basis expansions via stacked residual blocks with some structural interpretability through Fourier and polynomial constraints.
Transformer and state-space architectures~\cite{wu2021_autoformer,zhou2022_fedformer,lim2021_tft,gu2024_mamba} and pre-trained foundation models~\cite{das2024_timesfm,ansari2024_chronos} achieve strong forecasting accuracy, yet the underlying signal structure remains distributed across millions of parameters with no compact mathematical description.
We target this gap, by decomposing the signal into a small set of closed-form components whose parameters are directly readable.


\section{Methodology}
\label{sec:methodology}
We formulate the interpretable signal decomposition problem as finding a sparse set of parametric functions that best approximate the observed time series $\mathbf{y} \in \mathbb{R}^N$ over time $\mathbf{t}=[t_1,\dots,t_N]^\top \in \mathbb{R}^N$, where $t \in \mathbb{R}$ denotes the continuous time variable and $\psi_k(\mathbf{t};\,\theta_k) \in \mathbb{R}^N$ denotes component-wise evaluation over the grid.

\textbf{Problem Formulation:}
\label{sec:problem_formulation}
Let $\mathcal{A}$ be a set of atom families (e.g., Sine, Gaussian, Linear). 
Each family $a \in \mathcal{A}$ defines a parametric manifold $\mathcal{M}_a = \{ \psi(t;\,\theta) : \theta \in \Theta_a \}$, where $\theta$ collects the continuous parameters (frequency, amplitude, width, etc.).
The union of these manifolds forms our continuous dictionary $\mathcal{D} = \bigcup_{a \in \mathcal{A}} \mathcal{M}_a$.\\
Our goal is to find a set of $K$ atoms $\{\boldsymbol \psi_k\}_{k=1}^K$ and their parameters $\{\boldsymbol \theta_k\}_{k=1}^K$ to minimize the reconstruction error subject to a sparsity constraint. 
This can be formalized as an $\ell_0$ regularized nonlinear least squares problem:
\begin{equation}
    \min_{K, \{\boldsymbol\theta_k\}_{k=1}^K} \underbrace{ \frac{1}{2} \left\| \mathbf{y} - \sum_{k=1}^K \boldsymbol\psi_k(\mathbf{t}; \boldsymbol\theta_k) \right\|_2^2 }_{\text{Data Fidelity Term}} + \underbrace{ \lambda K }_{\text{Sparsity Penalty}}
    \label{eq:objective}
\end{equation}
Here, the objective function balances two competing goals: accuracy and simplicity.
The first term measures reconstruction error, driving parameters $\theta_k$ to capture signal morphology precisely. 
The second term penalizes model complexity, admitting a new atom only when it reduces the error by more than $\lambda$, which determines the model order automatically and prevents overfitting. 
The discrete analogue of this problem with a fixed dictionary is NP-hard~\cite{natarajan1995_sparse,davis1997_adaptive}, and the continuous formulation considered here remains non-convex with no known efficient global solver. 
CDP therefore approximates the solution via greedy forward selection with gradient-based refinement.
Instead of tuning $\lambda$ directly, we treat the model order $K$ as the sparsity control. By sweeping $K$ over a set budget during our evaluations, we trace the Pareto frontier of accuracy vs. sparsity, where each fixed-$K$ solution implicitly optimizes the objective for some $\lambda$.

\subsection{The Parametric Dictionary Library}
\label{sec:parametric_dictionary_library}
\begin{figure}[htbp]
  \begin{center}
    \includegraphics[width=0.9\columnwidth]{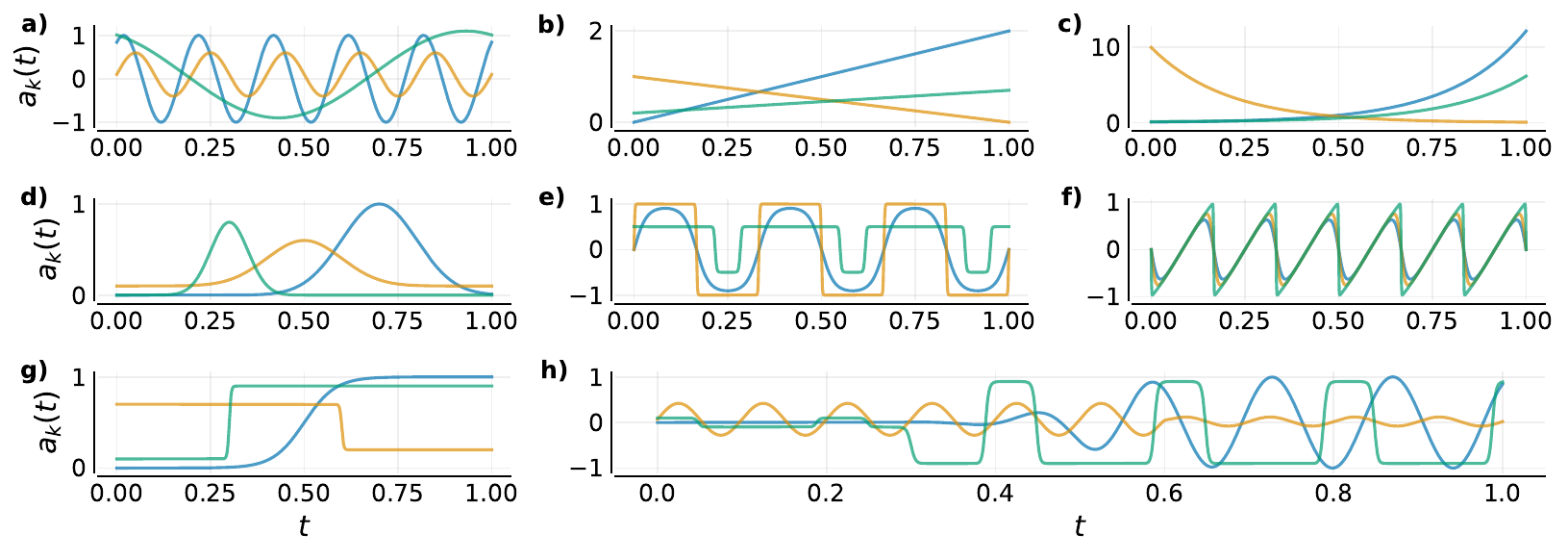}
    \caption{\textbf{Parametric Dictionary Library:} Example parameterizations of atom types: (a) SineAtom, (b) LinearAtom, (c) ExponentialAtom, (d) GaussianAtom, (e) SquareAtom, (f) SawtoothAtom, (g) StepAtom, (h) ModulatedAtom (for example $\texttt{SineAtom} \cdot \texttt{StepAtom}$).}
    \label{fig:all_atoms}
  \end{center}
\end{figure}%
The efficacy of the CDP framework relies on a diverse library of parametric continuous and discontinuous atoms categorized into periodic, trend, shape, and modulated families.
Every atom is built on a shared structure decomposing $\theta_k = (\phi_k, b_k)$ into atom specific parameters $\phi_k$ and a scalar bias $b_k \in \mathbb{R}$:
\begin{equation}
    \psi_k(t; \theta_k) = a_k(t; \phi_k) + b_k
\end{equation}
The bias $b_k$ enables individual components to adapt to localized mean offsets without requiring a separate trend atom.
For the discontinuous atoms, CDP offers differentiable approximations controlled by a sharpness scalar $\kappa > 0$, which is exponentially annealed during optimization, providing smooth gradients early and sharp waveform edges in the final epochs:
\begin{equation}
    \kappa(\text{epoch}) = \min\!\left(\kappa_{\text{end}},\; \kappa_{\text{start}} 
    \cdot e^{\gamma \cdot \text{epoch}}\right)
\end{equation}

\textbf{Trend and Shape Atoms}
\label{sec:trend_shape_atoms}
Global and transient structure is covered by four families (\autoref{fig:all_atoms}b,c,d,g). 
The LinearAtom and ExponentialAtom capture monotonic behavior, the GaussianAtom isolates localized transient events, and the StepAtom models abrupt level shifts. 
The ExponentialAtom is conditionally time-shifted based on the sign of the growth rate $r$, bounding the unscaled exponential to a maximum of $1$ and decoupling amplitude from rate to prevent gradient explosion~\cite{boyd2004_convexoptim}:
\begin{align}
    a_{\text{linear}}(t;\,\phi) &= m \cdot t \label{eq:linear}\\
    a_{\text{exp}}(t;\,\phi) &= A \cdot \exp\!\left( \begin{cases}
          r(t - T) & \text{if } r \ge 0 \\
          r \cdot t & \text{if } r < 0
       \end{cases}\right) \label{eq:exp}\\
    a_{\text{gauss}}(t;\,\phi) &= A \cdot \exp\!\left(-\tfrac{1}{2}
    \left(\tfrac{t - \mu}{\sigma}\right)^{\!2}\right) \label{eq:gauss}\\
    a_{\text{step}}(t;\,\phi) &= A \cdot \sigma\!\left( \tfrac{8\kappa}{T}
    (t - t_{\text{step}}) \right) \label{eq:step}
\end{align}
where $m$, $A$, $r$, $T$, $\mu$, $\sigma$, $t_{\text{step}}$ are the scalar entries of $\phi$.
The normalization factor $\frac{8}{T}$ in the StepAtom makes $\kappa$ scale-invariant across series of different lengths.\\
The ModulatedAtom (\autoref{fig:all_atoms}h) multiplies any carrier and modulator pair from the library, enabling composite structures such as damped oscillations or localized burst periodicities:
\begin{equation}
    a_{\text{mod}}(t;\,\phi_{\text{c}},\phi_{\text{m}}) = a_{\text{carrier}}(t;\,\phi_{\text{c}}) \cdot
    a_{\text{modulator}}(t;\,\phi_{\text{m}}) \label{eq:mod}
\end{equation}

\textbf{Periodic Atoms}
\label{sec:periodic_atoms}
Oscillatory behavior is anchored by the SineAtom. For non-sinusoidal periodicity, CDP provides differentiable approximations of two further waveform families (\autoref{fig:all_atoms}a,e,f).
The SquareAtom supports an adjustable duty cycle $\text{D} \in (0,1)$ that widens or narrows the rectangular pulses~\cite{yu1997_pwm}, and the SawtoothAtom captures asymmetric rise and fall patterns~\cite{noble1962_sawtoothinbio}:
\begin{align}
    a_{\text{sine}}(t;\,\phi) &= A \cdot \sin(2\pi f t + \varphi) \label{eq:sine}\\
    a_{\text{square}}(t;\,\phi) &= A \cdot \tanh\!\left(\tfrac{\kappa}{2}
    \cdot (\sin(2\pi f t + \varphi) - \cos(\pi D))\right) \label{eq:square}\\
    a_{\text{saw}}(t;\,\phi) &= A \cdot \left( \tanh(2\kappa \cos\varphi_1)
    \sin\varphi_1 + \tanh(2\kappa \cos\varphi_2) \sin\varphi_2 \right) \label{eq:saw}
\end{align}
where $\varphi_{1,2} = \tfrac{1}{4}(2\pi f t + \varphi \pm \pi)$ and $A, f, \varphi, D$ are the scalar entries of $\phi$.

\textbf{Theoretical Properties.}
The SineAtom manifold contains the $N \times N$ DCT-II basis as a special case and therefore spans $\mathbb{R}^N$, so the dictionary is complete (proof in \autoref{appendix:completeness}).
The remaining atom families enrich the dictionary with morphological structures that a pure sinusoidal basis cannot represent compactly.
Completeness alone establishes representability, not convergence or recovery.
In \autoref{appendix:woga} we therefore show that CDP is an instance of the weak orthogonal greedy algorithm~\cite{temlyakov2000_weakgreedy}.
In this framework, completeness guarantees that the atomic norm $|\mathbf{y}|_{\mathcal{A}_1(\mathcal{D})}$ is finite, which is exactly the hypothesis of an explicit residual decay bound.
The single condition of this bound is a per-step approximation ratio that can be computed exactly from every fitted model.
Verifying it on all 13 real-world datasets, the certified residual bound tracks the actual training error with a median ratio of $1.12$, and against a randomized reference search over the continuous dictionary the realized weakness parameter exceeds $0.8$ in $85\%$ of the greedy steps (\autoref{appendix:woga}).
In \autoref{appendix:gamma} we further prove that the annealed surrogate objectives for discontinuous atoms converge uniformly to their ideal sharp-limit counterparts at an explicit $O(\kappa^{-1/2})$ rate, which implies $\Gamma$-convergence, so the smoothing is a consistent approximation rather than a heuristic.
Formal global recovery and identifiability guarantees for mixed parametric families remain open for all methods in this class.
Existing guarantees~\cite{bhaskar2013_atomicnorm,tang2013_compressed,candes2014_superres} cover only the sinusoidal case, and our synthetic parameter-recovery experiments (\autoref{sec:synthetic_eval}) serve as their empirical counterpart.

\subsection{Greedy Forward Selection via Data-Driven Priors and Truncated Optimization}
\label{sec:greedy}
We iteratively solve the sub-problem of finding the optimal atomic component to approximate the current residual $\mathbf{r}_{k-1} = \mathbf{y} - \sum_{i=1}^{k-1}\psi_i(\mathbf{t};\,\theta_i) \in \mathbb{R}^N$ as shown in \autoref{fig:cdp_diagram}. 
At iteration $k$, we seek parameters $\theta^*$ such that:
\begin{equation}
    \theta^*_k = \arg\min_{a \in \mathcal{A},\;\theta \in \Theta_a} \|\mathbf{r}_{k-1} - \psi(\mathbf{t};\,\theta)\|_2^2
\end{equation}
The highly non-convex nature of this parametric landscape makes random initialization susceptible to poor local minima.
CDP therefore initializes candidates from a data-driven proposal distribution: periodic atoms start at dominant frequencies extracted via the Fast Fourier Transform (FFT) of the residual, while shape-based atoms are sampled from the residual's empirical variance and temporal support, with localization parameters bounded within the active time window and amplitudes scaled to the residual energy (\autoref{appendix:data_driven_priors}).

To isolate the best candidate efficiently, a diverse population of instantiated atoms is optimized in parallel for a fixed truncated budget, which shifts candidates toward promising basins of attraction without the cost of full convergence.
The candidate with the minimal reconstruction error is then added to the active dictionary and fully optimized before the next greedy iteration.
A global refinement step additionally performs a joint optimization over all active parameters $\Theta_k = (\theta_1, \dots, \theta_k)$, correcting suboptimal assignments from earlier iterations by accounting for correlations between selected components, with a configurable frequency from after every atom to periodically after a predefined batch.

All optimization stages use L-BFGS~\cite{liu1989_lbfgs}, whose second-order curvature information decouples dimensions with highly disparate sensitivities such as frequency versus amplitude.
To prevent physically invalid parameters, for example negative frequencies or vanishing bandwidths, the MSE objective is constrained with soft barrier penalties that smoothly force each parameter to its physically meaningful range (details in \autoref{appendix:limit_losses}).


\section{Experiments}
\label{sec:experiments}
\subsection{Experimental Setup}
\label{sec:experimental_setup}
We evaluate CDP on both synthetic and real-world datasets.
CDP's gradient-based optimization trades a higher one-time training cost for a closed-form model with efficient inference, and hardware details, runtimes, and hyperparameters are reported in \cref{appendix:runtime_eval,appendix:hyperparams}.
All aggregated experiments and figures report the Interquartile Mean (IQM) across random seeds with $95\%$ confidence intervals from percentile bootstrap, which is robust to outliers in small-sample evaluations~\cite{agarwal2021_rliable}.
All evaluations use 10 seeds, covering the stochastic methods of the main benchmark (CDP, CDP (Modulated), N-BEATS, NOMP, SFW, MCA, Analysis K-SVD) and the curriculum experiments. 
The deterministic baselines (MSTL, SSA, OMP, Prophet) are run once.

\textbf{Evaluation Datasets:}
\label{sec:eval_datasets}
As the true decomposition of real-world signals is typically unknown, we first constructed a custom synthesizer to generate signals with varying complexities, including harmonics, trends, damped oscillations, and non-stationarities, which allows exact ground-truth validation of parameter recovery.
For real-world validation, we evaluate on 13 series drawn from the standard LTSF benchmarks ETT~\cite{zhou2021_informer}, Electricity~\cite{zhou2021_informer,lai2018_modelingltsf}, Exchange~\cite{lai2018_modelingltsf}, PEMS~\cite{guo2019_astgcn}, Solar~\cite{lai2018_modelingltsf}, Traffic~\cite{lai2018_modelingltsf}, and Weather~\cite{wu2021_autoformer}, spanning diverse domains, sampling resolutions, and signal characteristics.
Although these benchmarks are multivariate, CDP is designed for univariate decomposition, and we follow established practice by selecting the first variate per dataset.

\textbf{Competitors:}
\label{sec:competitors}
We benchmarked CDP against methods spanning classical decomposition, sparse approximation, additive modeling, continuous pursuit, structured dictionary learning, and deep learning.
MSTL~\cite{bandara2025_mstl} and SSA~\cite{vautard1992_ssa} represent classical structural decomposition, OMP~\cite{pati1993_omp} with a discrete Fourier dictionary serves as a direct sparse spectral baseline, Prophet~\cite{taylor2018_prophet} as a widely used parametric additive model, and N-BEATS~\cite{oreshkin2019_nbeats} as a neural baseline whose residual block architecture is explicitly motivated by signal decomposition.
We further include the two closest continuous pursuit methods, Newtonized OMP (NOMP)~\cite{mamandipoor2016_nomp} and Sliding Frank-Wolfe (SFW)~\cite{denoyelle2019_sfw}, which refine atom parameters off the grid but operate on a single analytic family.
Finally, we add two structured-dictionary baselines.
Morphological component analysis (MCA)~\cite{starck2005_mca,bobin2007_mca} is instantiated with the same primitive families as CDP but with parameters frozen on discrete grids, which makes it a discretized ablation of our method, and Analysis K-SVD~\cite{rubinstein2013_analysisksvd} represents analysis-model dictionary learning.
Implementation details for all baselines are provided in \autoref{appendix:baselines}.

\subsection{Synthetic Evaluation and Parameter Extraction}
\label{sec:synthetic_eval}
\begin{figure}[htbp]
\begin{center}
\includegraphics[width=0.9\columnwidth]{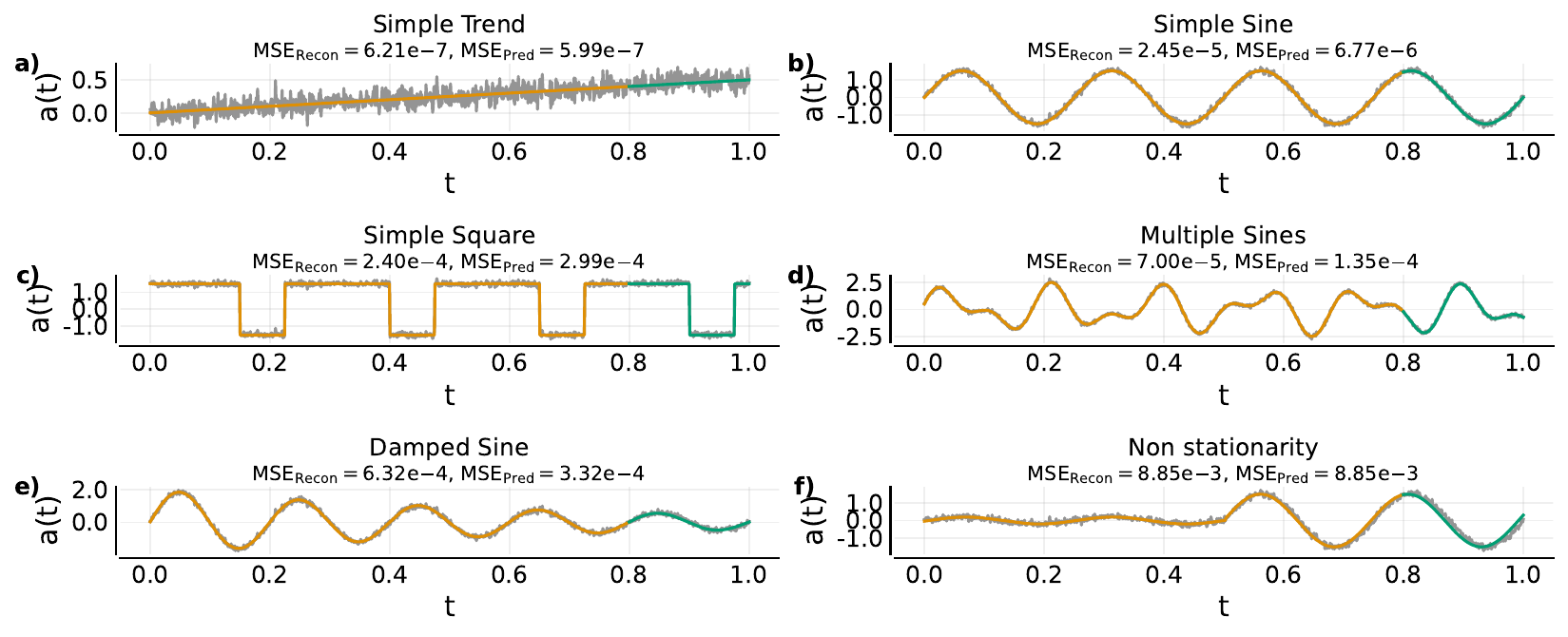}
\caption{\textbf{Parameter recovery on synthetic datasets:} CDP reconstructs the true mathematical components from noisy targets ($\sigma=0.1$): (a) linear trend, (b) sine, (c) square wave, (d) off-grid frequency superposition, (e) damped sine via a modulated atom, (f) non-stationary amplitude shift. Reconstruction and prediction MSE are given per panel.}
\label{fig:synthetic_dataset_evaluation}
\end{center}
\end{figure}%
As a basic sanity check, we test whether CDP identifies the true underlying components, all of which can be expressed by atoms in the library, rather than simply fitting their noisy sum. Both input and target signals are corrupted with additive Gaussian noise ($\sigma = 0.1$), while reconstruction quality is assessed against the clean ground truth.

We evaluate six signal structures of increasing complexity: a linear trend, a sine wave, a square wave, an off-grid frequency superposition, a damped sine wave, and a non-stationary amplitude shift.
Across all six, CDP recovers the generative parameters to high precision, as shown in \autoref{fig:synthetic_dataset_evaluation}. 
The recovered parameters deviate from the ground truth by less than $1\%$ in all stationary cases. 
For off-grid frequencies, where standard OMP produces a reconstruction MSE of $0.36$ due to spectral leakage, CDP achieves $7.00\mathrm{e}{-5}$. 
For the non-stationary amplitude shift, CDP places the transition point within $1\%$ of the true location using a single modulated atom, while OMP yields an MSE of $0.31$ and fails entirely on forecasting. 
The full parameter recovery equations for all six experiments are provided in \autoref{appendix:synthetic_eval}.

\subsection{Sparsity vs. Accuracy (Pareto Analysis)}
\label{sec:reconstruction_eval}
\begin{figure}[htbp]
\begin{center}
\includegraphics[width=0.9\columnwidth]{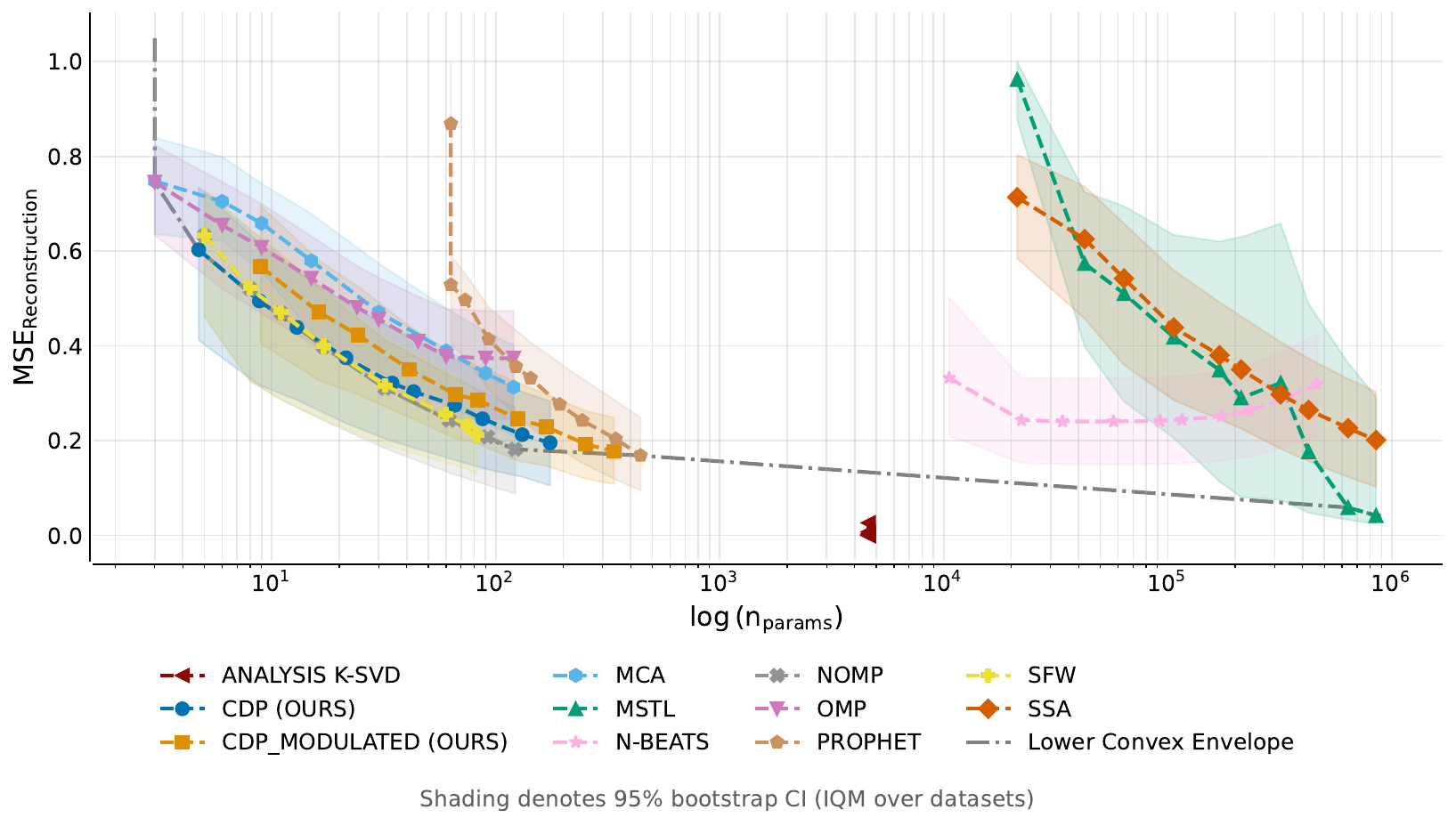}
\caption{\textbf{Pareto frontier of parameter count versus training reconstruction MSE:} CDP and CDP (Modulated) achieve the optimal trade-off in the sparse regime across all baselines. The lower convex envelope excludes Analysis K-SVD, whose near-zero training residual reflects in-sample denoising rather than a parametric decomposition (\autoref{appendix:baselines}).}
\label{fig:pareto_evaluation}
\end{center}
\end{figure}%
To evaluate the trade-off between the number of parameters and the reconstruction Mean Squared Error (MSE), we constructed a Pareto frontier across the real-world LTSF datasets (\autoref{fig:pareto_evaluation}). The individual evaluations of each dataset can be found in the appendix (\ref{appendix:individual_dataset}).

A core objective of CDP is achieving high reconstruction accuracy with minimal parameters.
Both CDP and CDP (Modulated) define the lower convex envelope at low parameter counts ($10^1$–$10^2$), meaning no baseline achieves a better accuracy-sparsity trade-off in this regime.
Because a single continuous atom can resolve a physical component exactly, CDP reduces error faster per added parameter than OMP, which requires multiple neighboring atoms to approximate off-grid frequencies.
High-parameter baselines such as MSTL and SSA only enter the plot beyond $10^4$ parameters and do not reach competitive error until $10^5$–$10^6$, while Prophet plateaus at a high MSE regardless of parameter count.
CDP thus occupies the efficient frontier in the sparse regime it targets while remaining fully parametrically transparent.
Out-of-sample forecasting results follow a similar pattern, with performance degrading only when non-stationary domain shifts occur between train and test sets (See \autoref{appendix:forecasting_eval} for full analysis).

\textbf{Comparison with continuous pursuit methods:}
The comparison with NOMP and SFW, which share CDP's off-grid refinement but not its mixed-morphology library, reveals two regimes.
In the sparse regime, CDP achieves the best median test error across the 13 datasets, for example $0.62$ against $1.05$ for NOMP and $0.84$ for SFW at $K{=}3$, while at larger budgets ($K \geq 20$) the single-family methods win through their exact update steps (\autoref{appendix:baselines}).
The morphology breakdown in \autoref{tab:morphology_breakdown} (analysis in \autoref{appendix:baselines}) separates the two effects: with a matched sinusoid-only library, CDP's optimizer is equivalent to the single-family baselines on strongly periodic data, and the full library adds coverage gains of $13$--$28\%$ on datasets with switching or spiky structure.
\begin{table}[htbp]
\centering
\caption{\textbf{Per-dataset training reconstruction loss at $K=5$} for the continuous and structured baselines, the CDP-sine ablation, and full CDP. The best value per dataset is bold, excluding Analysis K-SVD, whose near-zero training residual reflects in-sample denoising rather than a parametric decomposition (\autoref{appendix:baselines}).}
\label{tab:morphology_breakdown}
\small
\begin{tabular}{lcccccc}
\toprule
\textbf{Dataset} & \textbf{NOMP} & \textbf{SFW} & \textbf{MCA} & \textbf{AK-SVD} & \textbf{CDP-sine} & \textbf{CDP} \\ \midrule
Electricity & 0.562 & 0.567 & 0.622 & 0.007 & 0.593 & \textbf{0.514} \\
ETTh1 & \textbf{0.418} & 0.421 & 0.594 & 0.005 & 0.486 & 0.427 \\
ETTh2 & \textbf{0.539} & 0.545 & 0.760 & 0.004 & 0.692 & 0.545 \\
ETTm1 & \textbf{0.420} & 0.422 & 0.575 & 0.002 & 0.488 & 0.426 \\
ETTm2 & 0.537 & 0.542 & 0.740 & 0.002 & 0.691 & \textbf{0.491} \\
Exchange & \textbf{0.055} & 0.060 & 0.135 & 0.000 & 0.145 & 0.066 \\
PEMS03 & \textbf{0.196} & \textbf{0.196} & 0.384 & 0.001 & \textbf{0.196} & 0.201 \\
PEMS04 & 0.250 & 0.250 & 0.475 & 0.002 & 0.253 & \textbf{0.238} \\
PEMS07 & 0.429 & 0.429 & 0.590 & 0.002 & 0.436 & \textbf{0.313} \\
PEMS08 & \textbf{0.138} & \textbf{0.138} & 0.490 & 0.001 & \textbf{0.138} & 0.157 \\
Solar & \textbf{0.161} & 0.167 & 0.187 & 0.001 & 0.167 & 0.182 \\
Traffic & \textbf{0.550} & \textbf{0.550} & 0.712 & 0.003 & \textbf{0.550} & 0.579 \\
Weather & 0.589 & 0.595 & 0.791 & 0.000 & 0.599 & \textbf{0.496} \\ \bottomrule
\end{tabular}
\end{table}
The matched-family MCA ablation isolates the value of continuous parameter optimization itself, as reconstruction with identical primitive families but grid-frozen parameters is worse at every budget (median $0.59$ against $0.43$ at $K{=}5$).
Analysis K-SVD illustrates the analysis-model alternative, which achieves a near-zero in-sample residual but offers unstable extrapolation and no closed-form atoms.

\subsection{Interpretability in Real-World Data}
\label{sec:interpretability_eval}
\begin{figure}[htbp]
\begin{center}
\includegraphics[width=0.9\columnwidth]{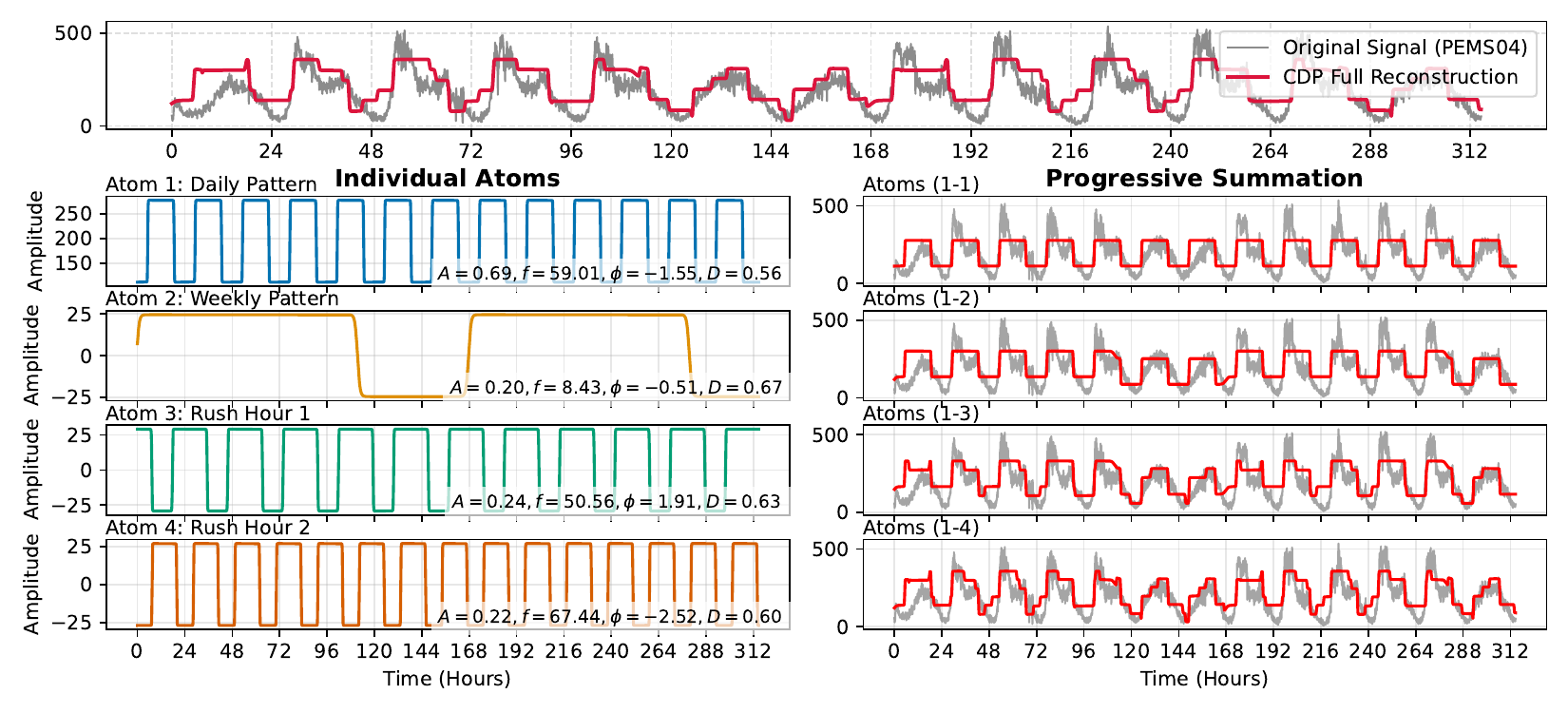}
\caption{\textbf{Signal decomposition of the PEMS04 traffic dataset:} Top: the original signal overlaid with the full CDP reconstruction. Left: the individual extracted atoms, capturing daily cycles, weekly patterns, and localized rush-hour modulations. Right: their progressive summation toward the target signal.}
\label{fig:interpretability_evaluation}
\end{center}
\end{figure}%
We qualitatively assessed the physical meaning of the extracted components on three example datasets.
Because each atom's frequency maps directly to a physical period, the extracted components can be read in real time units.
On PEMS07, CDP extracts a square-wave daily pattern whose duty cycle of 0.76 matches the split between high daytime and low nighttime traffic.
On ETTm1, the dominant component is the 24-hour temperature cycle, recovered without any domain-specific prior.
For PEMS04, shown in \autoref{fig:interpretability_evaluation}, the decomposition separates a daily pattern, a weekly pattern capturing weekend drops, and localized rush-hour peaks, each corresponding to a single atom.
To quantify how each method distributes explanatory weight across its components, and whether individual atoms capture distinct structures rather than spreading their contribution, we conduct an additional SHAP analysis~\cite{lundberg2017_shapvalues} in \autoref{appendix:shap_eval}.

\subsection{Atom-Based Curriculum Learning}
\label{sec:curriculum_eval}

\begin{figure}[htbp]
  \begin{center}
    \includegraphics[width=0.9\columnwidth]{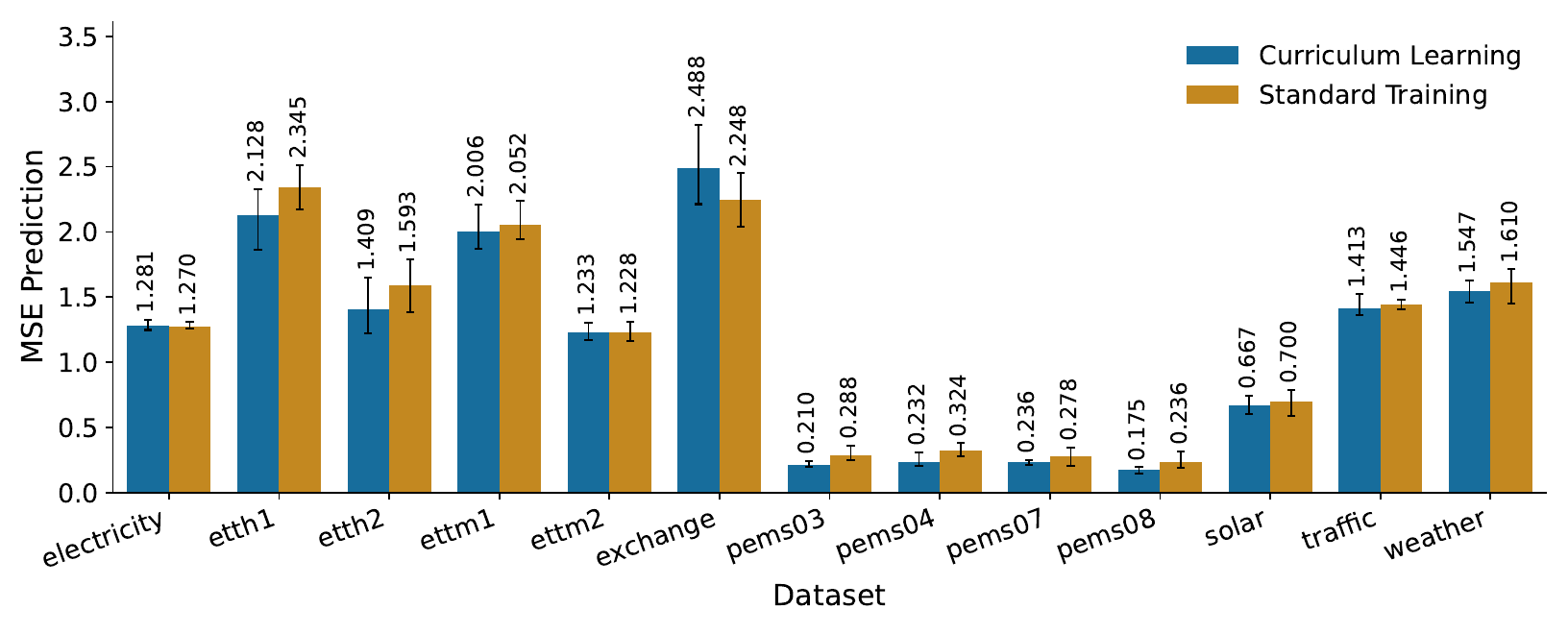}
    \caption{\textbf{Forecasting MSE for curriculum learning versus standard end-to-end training} across all 13 real-world LTSF datasets (IQM over 10 seeds, whiskers show 95\% bootstrap confidence intervals).}
    \label{fig:curriculum_evaluation}
  \end{center}
\end{figure}%

A key motivation for obtaining a structured parametric decomposition is that the recovered atoms encode the signal's components in a form that can directly benefit downstream learning.
Following Bengio et al.~\cite{bengio2009_curriculum}, we train a vanilla S4D state space sequence model~\cite{gu2022_s4d} by introducing atoms one at a time, blending each into the target via a sigmoid-shaped curriculum weight, before a final fine-tuning stage on the normalized real signal.
The full procedure is described in \autoref{appendix:curriculum_learning}.

All curricula are evaluated under a single protocol on the 13 real-world LTSF datasets.
The S4D model, the staging schedule, and the fine-tuning phase are held fixed, and only the curriculum targets vary, covering CDP atoms in greedy coarse-to-fine order, the components of the additive baselines (SSA, MSTL, OMP, Prophet, NOMP, SFW), CDP atoms in reversed fine-to-coarse order, a decomposition-free low-pass smoothing of the raw signal, and standard end-to-end training.
Every condition uses the stored seed-0 decompositions at an atom budget of 10, and every reported value is the Interquartile Mean over 10 training seeds with 95\% percentile bootstrap confidence intervals.

Across the 13 datasets, staging the CDP atoms in greedy coarse-to-fine order reduces the out-of-sample forecasting error relative to standard end-to-end training on 10 of 13 datasets, with relative reductions of 25 to 29 percent on PEMS03, PEMS04, and PEMS08 (\autoref{fig:curriculum_evaluation}).
The confidence intervals separate fully on PEMS03, and the curriculum is not significantly worse than standard training on any dataset.
The reversed fine-to-coarse ordering performs worse than the greedy coarse-to-fine ordering on 11 of 13 datasets, so the ordering contributes a substantial part of the gain.

Comparing curricula built from all baseline decompositions (\autoref{tab:curriculum_results}), CDP atoms are the best target on 8 of 13 datasets, and the benefit tracks the forecast fidelity of the underlying decomposition.
On Exchange and Weather, where CDP extrapolates poorly (\autoref{appendix:forecasting_eval}), the curriculum advantage shrinks. The diverging SFW curriculum on Exchange is the extreme case: it selects nearly identical atoms whose large weights largely offset each other (\autoref{appendix:shap_eval}), so partial sums grow far larger than the signal itself.
CDP atoms are thus useful structured training targets precisely where CDP decomposes the signal accurately.

\begin{table}[htbp]
\centering
\caption{\textbf{Curriculum evaluation on all datasets.} Out-of-sample forecast MSE of the S4D model for every curriculum target (IQM over 10 seeds). The best value per dataset is bold.}
\label{tab:curriculum_results}
\scriptsize
\setlength{\tabcolsep}{3.5pt}
\begin{tabular}{lcccccccccc}
\toprule
\textbf{Dataset} & \textbf{CDP} & \textbf{CDP (rev.)} & \textbf{SSA} & \textbf{MSTL} & \textbf{OMP} & \textbf{Prophet} & \textbf{NOMP} & \textbf{SFW} & \textbf{Smooth.} & \textbf{Standard} \\ \midrule
Electricity & 1.281 & 1.277 & 1.286 & 1.295 & 1.269 & 1.273 & \textbf{1.254} & 1.269 & 1.261 & 1.270 \\
ETTh1 & \textbf{2.128} & 2.244 & 2.272 & 2.285 & 2.216 & 2.199 & 2.222 & 2.241 & 2.283 & 2.345 \\
ETTh2 & 1.409 & 1.675 & 1.566 & 1.449 & 1.503 & 1.496 & 1.497 & \textbf{1.397} & 1.492 & 1.593 \\
ETTm1 & \textbf{2.006} & 2.105 & 2.162 & 2.158 & 2.100 & 2.177 & 2.065 & 2.136 & 2.183 & 2.052 \\
ETTm2 & 1.233 & 1.272 & 1.233 & \textbf{1.218} & 1.227 & 1.254 & 1.220 & 1.285 & 1.236 & 1.228 \\
Exchange & 2.488 & 2.340 & 2.265 & 2.260 & 2.062 & 2.335 & 2.323 & $9.8{\times}10^{4}$ & \textbf{1.824} & 2.248 \\
PEMS03 & \textbf{0.210} & 0.251 & 0.298 & 0.235 & 0.310 & 0.292 & 0.291 & 0.296 & 0.453 & 0.288 \\
PEMS04 & \textbf{0.232} & 0.264 & 0.298 & 0.583 & 0.247 & 0.314 & 0.286 & 0.293 & 0.348 & 0.324 \\
PEMS07 & \textbf{0.236} & 0.248 & 0.489 & 0.401 & 0.379 & 0.481 & 0.418 & 0.446 & 0.645 & 0.278 \\
PEMS08 & \textbf{0.175} & 0.304 & 0.301 & 0.276 & 0.269 & 0.266 & 0.244 & 0.261 & 0.402 & 0.236 \\
Solar & \textbf{0.667} & 0.678 & 0.768 & 0.750 & 0.694 & 0.699 & 0.724 & 6.451 & 0.799 & 0.700 \\
Traffic & \textbf{1.413} & 1.574 & 1.434 & 1.582 & 1.495 & 1.463 & 1.428 & 1.443 & 1.610 & 1.446 \\
Weather & 1.547 & 1.647 & 1.566 & 1.554 & \textbf{1.322} & 1.501 & 1.456 & 1.500 & 1.414 & 1.610 \\ \bottomrule
\end{tabular}
\end{table}


\section{Limitations}
\label{sec:limitations}
CDP rests on a non-convex greedy optimization.
The convergence result in \autoref{appendix:woga} is conditional on a per-step approximation ratio that we measure rather than prove, formal identifiability guarantees for mixed parametric families remain open, and the high stability of the recovered atoms across seeds, time windows, and hyperparameters (\autoref{appendix:robustness}) is an empirical observation rather than a guarantee.
The method further depends on its predefined atom library, and for out-of-library structure the fit degrades toward a best-in-library approximation, so atoms extracted from real-world signals, where no ground-truth decomposition exists, should not be over-interpreted as true physical components.
Finally, fitting is more expensive than classical grid-based greedy methods (\autoref{appendix:runtime_eval}), although the cost is a one-time investment and inference on the closed-form model is the cheapest of all evaluated methods, and CDP is currently limited to univariate signals, with a multivariate extension that shares atoms across channels left to future work.


\section{Conclusion}
\label{sec:conclusion}
This paper presents Continuous Dictionary Pursuit (CDP), a greedy iterative framework that removes the grid discretization error and the resulting spectral leakage of fixed-dictionary sparse coding for atom families contained in its library, by optimizing directly over a continuous parametric manifold that includes differentiable approximations of discontinuous shapes.
On real-world benchmarks, CDP occupies a favorable position on the accuracy-sparsity Pareto frontier with orders of magnitude fewer parameters than deep learning baselines, and the extracted atoms serve as structured curriculum targets that reduce the out-of-sample forecasting error of neural sequence models on most evaluated datasets.
The full implementation is publicly available to support reproducibility and further research.

CDP represents a meaningful step toward machine learning models that are not only accurate but also fully transparent, producing compact closed-form decompositions whose components can be interpreted directly.

\section*{Acknowledgements}
This work has been supported by the European Union under grant agreement no.~101311180 (DEEP-HEALTH), the MWK of Lower Saxony within Hybrint (VWZN4219) and LCIS (VWZN4704), and the DFG under Germany's Excellence Strategy within the Clusters of Excellence PhoenixD (EXC~2122) and Quantum Frontiers (EXC~2123).

\bibliographystyle{splncs04}
\bibliography{references}

@inproceedings{zhou2022_fedformer,
    title={Fedformer: Frequency enhanced decomposed transformer for long-term series forecasting},
    author={Tian Zhou and Ziqing Ma and Qingsong Wen and Xue Wang and Liang Sun and Rong Jin},
    booktitle={International conference on machine learning},
    pages={27268--27286},
    year={2022},
    organization={PMLR}
}

@misc{forootani2024_climateawaredeepneural,
    title={Climate Aware Deep Neural Networks (CADNN) for Wind Power Simulation}, 
    author={Ali Forootani and Danial Esmaeili Aliabadi and Daniela Thraen},
    year={2024},
    eprint={2412.12160},
    archivePrefix={arXiv},
    primaryClass={cs.LG},
}

@article{mustafa2025_energyforecasting,
    title = {Forecasting Energy Demand and Generation Using Time Series Models: A Comparative Analysis of Classical, Grey, Fuzzy, and Intelligent Approaches},
    journal = {Franklin Open},
    pages = {100350},
    year = {2025},
    author = {Anas Thamer Mustafa and Omar Sharaf Al-Deen Al-Yozbaky},
}

@book{tsay2005_financeanalysis,
  title={Analysis of financial time series},
  author={Tsay, Ruey S},
  year={2005},
  publisher={John wiley \& sons}
}

@inproceedings{chakraborti2007_financialoverview,
  title={Financial time-series analysis: A brief overview},
  author={Chakraborti, A and Patriarca, M and Santhanam, MS},
  booktitle={Econophysics of Markets and Business Networks: Proceedings of the Econophys-Kolkata III},
  pages={51--67},
  year={2007},
  organization={Springer}
}

@article{farahani2025_manufacturingsystems,
    title = {Time-series forecasting in smart manufacturing systems: An experimental evaluation of the state-of-the-art algorithms},
    author = {Mojtaba A. Farahani and Fadi El Kalach and Austin Harper and M.R. McCormick and Ramy Harik and Thorsten Wuest},
    journal = {Robotics and Computer-Integrated Manufacturing},
    volume = {95},
    pages = {103010},
    year = {2025},
}

@article{schaller2026_s4dbio,
    title = {Audio monitoring of bone cement disintegration in pulsating fluid jet surgery under laboratory conditions},
    journal = {Results in Engineering},
    volume = {29},
    pages = {108762},
    year = {2026},
    issn = {2590-1230},
    author = {Melanie Schaller and Sergej Hloch and Akash Nag and Dagmar Klichová and Frank Pude and Michal Zeleňák and Nick Janssen and Bodo Rosenhahn},
}

@INPROCEEDINGS{sreenivas2021_arrhythmiacnn,
  author={Vandith Sreenivas, K and Ganesan, M and Lavanya, R},
  booktitle={2021 Seventh International conference on Bio Signals, Images, and Instrumentation (ICBSII)}, 
  title={Classification of Arrhythmia in Time Series ECG Signals Using Image Encoding And Convolutional Neural Networks}, 
  year={2021},
  pages={1-6},
}

@InProceedings{bui2018_cardiocasculardiseases,
    author="Bui, C. and Pham, N. and Vo, A. and Tran, A. and Nguyen, A. and Le, T.", 
    title="Time Series Forecasting for Healthcare Diagnosis and Prognostics with the Focus on Cardiovascular Diseases",
    booktitle="6th International Conference on the Development of Biomedical Engineering in Vietnam (BME6) ",
    year="2018",
    publisher="Springer Singapore",
    address="Singapore",
    pages="809--818",
    isbn="978-981-10-4361-1"
}

@article{rios1996_typhoidfever,
  title={[Time series in the epidemiology of typhoid fever in Spain].},
  author={M R{\'i}os and J. M. Garc{\'i}a and Marta Cubedo and D. P{\'e}rez},
  journal={Medicina clinica},
  year={1996},
  volume={106 18},
  pages={686-9},
}

@article{rudin2019_blackbox,
  title={Stop explaining black box machine learning models for high stakes decisions and use interpretable models instead},
  author={Rudin, Cynthia},
  journal={Nature machine intelligence},
  volume={1},
  number={5},
  pages={206--215},
  year={2019},
  publisher={Nature Publishing Group UK London}
}

@article{doshi2017_towardsinterpretable,
  title={Towards a rigorous science of interpretable machine learning},
  author={Doshi-Velez, Finale and Kim, Been},
  journal={arXiv preprint arXiv:1702.08608},
  year={2017}
}

@article{goodman2017_euregulations,
  title={European Union regulations on algorithmic decision-making and a “right to explanation”},
  author={Goodman, Bryce and Flaxman, Seth},
  journal={AI magazine},
  volume={38},
  number={3},
  pages={50--57},
  year={2017}
}

@article{cleveland1990_stl,
  title={{STL}: A seasonal-trend decomposition procedure based on loess},
  author={Cleveland, Robert B and Cleveland, William S and McRae, Jean E and Terpenning, Irma},
  journal={Journal of Official Statistics},
  volume={6},
  number={1},
  pages={3--73},
  year={1990}
}

@article{mallat1993_matchingpursuit,
  title={Matching pursuits with time-frequency dictionaries},
  author={Mallat, St{\'e}phane G and Zhang, Zhifeng},
  journal={IEEE Transactions on signal processing},
  volume={41},
  number={12},
  pages={3397--3415},
  year={1993},
  publisher={IEEE}
}

@inbook{pezeshki2015_gridmismatch,
    author = {Pezeshki, Ali and Chi, Yuejie and Scharf, Louis and Chong, Edwin},
    year = {2015},
    month = {01},
    pages = {75-95},
    title = {Compressed Sensing, Sparse Inversion, and Model Mismatch},
    isbn = {978-3-319-16041-2},
    journal = {Applied and Numerical Harmonic Analysis},
    publisher = {Springer}
}

@article{da2021_clinicalinterpretability,
  title={Using interpretability approaches to update “black-box” clinical prediction models: an external validation study in nephrology},
  author={Da Cruz, Harry Freitas and Pfahringer, Boris and Martensen, Tom and Schneider, Frederic and Meyer, Alexander and B{\"o}ttinger, Erwin and Schapranow, Matthieu-P},
  journal={Artificial Intelligence in Medicine},
  volume={111},
  pages={101982},
  year={2021},
  publisher={Elsevier}
}

@article{singh2020_medicalimageexplainable,
  title={Explainable deep learning models in medical image analysis},
  author={Singh, Amitojdeep and Sengupta, Sourya and Lakshminarayanan, Vasudevan},
  journal={Journal of imaging},
  volume={6},
  number={6},
  pages={52},
  year={2020},
  publisher={MDPI}
}

@inproceedings{jing2022_inaction,
  title={Inaction: Interpretable action decision making for autonomous driving},
  author={Jing, Taotao and Xia, Haifeng and Tian, Renran and Ding, Haoran and Luo, Xiao and Domeyer, Joshua and Sherony, Rini and Ding, Zhengming},
  booktitle={European Conference on Computer Vision},
  pages={370--387},
  year={2022},
  organization={Springer}
}

@inproceedings{yan2023_selfinterpretable,
  title={Self-interpretable time series prediction with counterfactual explanations},
  author={Yan, Jingquan and Wang, Hao},
  booktitle={International Conference on Machine Learning},
  pages={39110--39125},
  year={2023},
  organization={PMLR}
}

@article{ozkurt2024_interpretablechaos,
  title={Interpretable AI analysis of chaos systems distribution in time series data from industrial robotics.},
  author={{\"O}zkurt, Cem},
  journal={Turkish Journal of Engineering (TUJE)},
  volume={8},
  number={4},
  year={2024}
}

@inproceedings{lundberg2017_shapvalues,
  title={A unified approach to interpreting model predictions},
  author={Lundberg, Scott M and Lee, Su-In},
  booktitle={Advances in Neural Information Processing Systems},
  volume={30},
  year={2017}
}

@article{taylor2018_prophet,
  title={Forecasting at scale},
  author={Taylor, Sean J and Letham, Benjamin},
  journal={The American Statistician},
  volume={72},
  number={1},
  pages={37--45},
  year={2018},
  publisher={Taylor \& Francis}
}

@inproceedings{pati1993_omp,
  title={Orthogonal matching pursuit: Recursive function approximation with applications to wavelet decomposition},
  author={Pati, Yagyensh Chandra and Rezaiifar, Ramin and Krishnaprasad, Perinkulam Sambamurthy},
  booktitle={Proceedings of 27th Asilomar conference on signals, systems and computers},
  pages={40--44},
  year={1993},
  organization={IEEE}
}

@inproceedings{oreshkin2019_nbeats,
  title={{N-BEATS}: Neural basis expansion analysis for interpretable time series forecasting},
  author={Oreshkin, Boris N and Carpov, Dmitri and Chapados, Nicolas and Bengio, Yoshua},
  booktitle={International Conference on Learning Representations},
  year={2020}
}

@article{vautard1992_ssa,
  title={Singular-spectrum analysis: A toolkit for short, noisy chaotic signals},
  author={Vautard, Robert and Yiou, Pascal and Ghil, Michael},
  journal={Physica D: Nonlinear Phenomena},
  volume={58},
  number={1-4},
  pages={95--126},
  year={1992},
  publisher={Elsevier}
}

@book{box1976_arima,
  title={Time Series Analysis: Forecasting and Control},
  author={Box, George E. P. and Jenkins, Gwilym M.},
  publisher={Holden-Day},
  address={San Francisco},
  year={1976}
}

@inproceedings{guo2019_astgcn,
  title={Attention Based Spatial-Temporal Graph Convolutional Networks for Traffic Flow Forecasting},
  author={Guo, Shengnan and Lin, Youfang and Feng, Ning and Song, Chao and Wan, Huaiyu},
  booktitle={Proceedings of the AAAI Conference on Artificial Intelligence},
  volume={33},
  pages={922--929},
  year={2019}
}

@article{bandara2025_mstl,
  title={MSTL: A seasonal-trend decomposition algorithm for time series with multiple seasonal patterns},
  author={Bandara, Kasun and Hyndman, Rob J and Bergmeir, Christoph},
  journal={International Journal of Operational Research},
  volume={52},
  number={1},
  pages={79--98},
  year={2025},
  publisher={Inderscience Publishers (IEL)}
}

@article{zhang2003_arimaforecasting,
  title={Time series forecasting using a hybrid ARIMA and neural network model},
  author={Zhang, G Peter},
  journal={Neurocomputing},
  volume={50},
  pages={159--175},
  year={2003},
  publisher={Elsevier}
}

@article{schmid2010_dmd,
  title={Dynamic mode decomposition of numerical and experimental data},
  author={Schmid, Peter J},
  journal={Journal of fluid mechanics},
  volume={656},
  pages={5--28},
  year={2010},
  publisher={Cambridge University Press}
}

@article{duarte2013_spectralcompressivesensing,
  title={Spectral compressive sensing},
  author={Duarte, Marco F and Baraniuk, Richard G},
  journal={Applied and Computational Harmonic Analysis},
  volume={35},
  number={1},
  pages={111--129},
  year={2013},
  publisher={Elsevier}
}

@article{lim2021_tft,
  title={Temporal fusion transformers for interpretable multi-horizon time series forecasting},
  author={Lim, Bryan and Ar{\i}k, Sercan {\"O} and Loeff, Nicolas and Pfister, Tomas},
  journal={International journal of forecasting},
  volume={37},
  number={4},
  pages={1748--1764},
  year={2021},
  publisher={Elsevier}
}

@inproceedings{wu2021_autoformer,
    title={Autoformer: Decomposition transformers with auto-correlation for long-term series forecasting},
    author={Wu, Haixu and Xu, Jiehui and Wang, Jianmin and Long, Mingsheng},
    booktitle={Advances in Neural Information Processing Systems},
    volume={34},
    pages={22419--22430},
    year={2021}
}

@book{boyd2004_convexoptim,
  title={Convex optimization},
  author={Boyd, Stephen and Vandenberghe, Lieven},
  year={2004},
  publisher={Cambridge university press}
}

@article{vstrumbelj2014_explainingfeatures,
  title={Explaining prediction models and individual predictions with feature contributions},
  author={{\v{S}}trumbelj, Erik and Kononenko, Igor},
  journal={Knowledge and information systems},
  volume={41},
  number={3},
  pages={647--665},
  year={2014},
  publisher={Springer}
}

@article{liu1989_lbfgs,
  title={On the limited memory BFGS method for large scale optimization},
  author={Liu, Dong C and Nocedal, Jorge},
  journal={Mathematical programming},
  volume={45},
  number={1},
  pages={503--528},
  year={1989},
  publisher={Springer}
}

@article{aharon2006_ksvd,
  title={K-SVD: An algorithm for designing overcomplete dictionaries for sparse representation},
  author={Aharon, Michal and Elad, Michael and Bruckstein, Alfred},
  journal={IEEE Transactions on signal processing},
  volume={54},
  number={11},
  pages={4311--4322},
  year={2006},
  publisher={IEEE}
}

@article{bhaskar2013_atomicnorm,
  title={Atomic norm denoising with applications to line spectral estimation},
  author={Bhaskar, Badri Narayan and Tang, Gongguo and Recht, Benjamin},
  journal={IEEE Transactions on Signal Processing},
  volume={61},
  number={23},
  pages={5987--5999},
  year={2013},
  publisher={IEEE}
}

@article{tang2013_compressed,
  title={Compressed sensing off the grid},
  author={Tang, Gongguo and Bhaskar, Badri Narayan and Shah, Parikshit and Recht, Benjamin},
  journal={IEEE transactions on information theory},
  volume={59},
  number={11},
  pages={7465--7490},
  year={2013},
  publisher={IEEE}
}

@inproceedings{gu2024_mamba,
  title={Mamba: Linear-time sequence modeling with selective state spaces},
  author={Gu, Albert and Dao, Tri},
  booktitle={First conference on language modeling},
  year={2024}
}

@inproceedings{das2024_timesfm,
  title={A decoder-only foundation model for time-series forecasting},
  author={Das, Abhimanyu and Kong, Weihao and Sen, Rajat and Zhou, Yichen},
  booktitle={Forty-first international conference on machine learning},
  year={2024}
}

@article{ansari2024_chronos,
  title={Chronos: Learning the language of time series},
  author={Ansari, Abdul Fatir and Stella, Lorenzo and Turkmen, Caner and Zhang, Xiyuan and Mercado, Pedro and Shen, Huibin and Shchur, Oleksandr and Rangapuram, Syama Sundar and Arango, Sebastian Pineda and Kapoor, Shubham and others},
  journal={Transactions on Machine Learning Research},
  year={2024}
}

@article{noble1962_sawtoothinbio,
  title={A modification of the {Hodgkin--Huxley} equations applicable to Purkinje fibre action and pacemaker potentials},
  author={Noble, Denis},
  journal={The Journal of physiology},
  volume={160},
  number={2},
  pages={317},
  year={1962}
}

@inproceedings{yu1997_pwm,
  title={A review of three PWM techniques},
  author={Yu, Zhenyu and Mohammed, Arefeen and Panahi, Issa},
  booktitle={Proceedings of the 1997 American Control Conference (Cat. No. 97CH36041)},
  volume={1},
  pages={257--261},
  year={1997},
  organization={IEEE}
}

@inproceedings{zhou2021_informer,
  title={Informer: Beyond efficient transformer for long sequence time-series forecasting},
  author={Zhou, Haoyi and Zhang, Shanghang and Peng, Jieqi and Zhang, Shuai and Li, Jianxin and Xiong, Hui and Zhang, Wancai},
  booktitle={Proceedings of the AAAI conference on artificial intelligence},
  volume={35},
  pages={11106--11115},
  year={2021}
}

@inproceedings{lai2018_modelingltsf,
  title={Modeling long-and short-term temporal patterns with deep neural networks},
  author={Lai, Guokun and Chang, Wei-Cheng and Yang, Yiming and Liu, Hanxiao},
  booktitle={The 41st international ACM SIGIR conference on research \& development in information retrieval},
  pages={95--104},
  year={2018}
}

@inproceedings{bengio2009_curriculum,
  title={Curriculum learning},
  author={Bengio, Yoshua and Louradour, J{\'e}r{\^o}me and Collobert, Ronan and Weston, Jason},
  booktitle={Proceedings of the 26th annual international conference on machine learning},
  pages={41--48},
  year={2009}
}

@inproceedings{gu2022_s4d,
  title={On the parameterization and initialization of diagonal state space models},
  author={Gu, Albert and Goel, Karan and Gupta, Ankit and R{\'e}, Christopher},
  booktitle={Advances in Neural Information Processing Systems},
  volume={35},
  pages={35971--35983},
  year={2022}
}

@article{natarajan1995_sparse,
  title={Sparse approximate solutions to linear systems},
  author={Natarajan, Balas Kausik},
  journal={SIAM journal on computing},
  volume={24},
  number={2},
  pages={227--234},
  year={1995},
  publisher={SIAM}
}

@article{davis1997_adaptive,
  title={Adaptive greedy approximations},
  author={Davis, Geoff and Mallat, Stephane and Avellaneda, Marco},
  journal={Constructive approximation},
  volume={13},
  number={1},
  pages={57--98},
  year={1997},
  publisher={Springer}
}

@book{rao2014_dct,
  title={Discrete cosine transform: algorithms, advantages, applications},
  author={Rao, K Ramamohan and Yip, Ping},
  year={2014},
  publisher={Academic press}
}

@inproceedings{agarwal2021_rliable,
  title={Deep reinforcement learning at the edge of the statistical precipice},
  author={Agarwal, Rishabh and Schwarzer, Max and Castro, Pablo Samuel and Courville, Aaron C and Bellemare, Marc},
  booktitle={Advances in Neural Information Processing Systems},
  volume={34},
  pages={29304--29320},
  year={2021}
}

@book{dumitrescu2018_dictionarylearning,
  title     = {Dictionary Learning Algorithms and Applications},
  author    = {Dumitrescu, Bogdan and Irofti, Paul},
  year      = {2018},
  publisher = {Springer},
  address   = {Cham}
}

@article{rubinstein2013_analysisksvd,
  title   = {Analysis {K-SVD}: A Dictionary-Learning Algorithm for the Analysis Sparse Model},
  author  = {Rubinstein, Ron and Peleg, Tomer and Elad, Michael},
  journal = {IEEE Transactions on Signal Processing},
  volume  = {61},
  number  = {3},
  pages   = {661--677},
  year    = {2013}
}

@inproceedings{qiu2012_domainadaptive,
  title     = {Domain Adaptive Dictionary Learning},
  author    = {Qiu, Qiang and Patel, Vishal M. and Turaga, Pavan and Chellappa, Rama},
  booktitle = {European Conference on Computer Vision (ECCV)},
  pages     = {631--645},
  year      = {2012},
  publisher = {Springer}
}

@article{tasissa2023_kdeepsimplex,
  title   = {K-Deep Simplex: Manifold Learning via Local Dictionaries},
  author  = {Tasissa, Abiy and Tankala, Pranay and Murphy, James M. and Ba, Demba},
  journal = {IEEE Transactions on Signal Processing},
  volume  = {71},
  pages   = {3741--3754},
  year    = {2023}
}

@article{starck2005_mca,
  title   = {Image Decomposition via the Combination of Sparse Representations and a Variational Approach},
  author  = {Starck, Jean-Luc and Elad, Michael and Donoho, David L.},
  journal = {IEEE Transactions on Image Processing},
  volume  = {14},
  number  = {10},
  pages   = {1570--1582},
  year    = {2005}
}

@article{bobin2007_mca,
  title   = {Morphological Component Analysis: An Adaptive Thresholding Strategy},
  author  = {Bobin, J{\'e}r{\^o}me and Starck, Jean-Luc and Fadili, Jalal M. and Moudden, Yassir and Donoho, David L.},
  journal = {IEEE Transactions on Image Processing},
  volume  = {16},
  number  = {11},
  pages   = {2675--2681},
  year    = {2007}
}

@article{mamandipoor2016_nomp,
  title   = {Newtonized Orthogonal Matching Pursuit: Frequency Estimation Over the Continuum},
  author  = {Mamandipoor, Babak and Ramasamy, Dinesh and Madhow, Upamanyu},
  journal = {IEEE Transactions on Signal Processing},
  volume  = {64},
  number  = {19},
  pages   = {5066--5081},
  year    = {2016}
}

@article{denoyelle2019_sfw,
  title   = {The Sliding {Frank-Wolfe} Algorithm and its Application to Super-Resolution Microscopy},
  author  = {Denoyelle, Quentin and Duval, Vincent and Peyr{\'e}, Gabriel and Soubies, Emmanuel},
  journal = {Inverse Problems},
  volume  = {36},
  number  = {1},
  pages   = {014001},
  year    = {2020}
}

@article{candes2014_superres,
  title   = {Towards a Mathematical Theory of Super-Resolution},
  author  = {Cand{\`e}s, Emmanuel J. and Fernandez-Granda, Carlos},
  journal = {Communications on Pure and Applied Mathematics},
  volume  = {67},
  number  = {6},
  pages   = {906--956},
  year    = {2014}
}

@article{temlyakov2000_weakgreedy,
  title   = {Weak Greedy Algorithms},
  author  = {Temlyakov, Vladimir N.},
  journal = {Advances in Computational Mathematics},
  volume  = {12},
  number  = {2--3},
  pages   = {213--227},
  year    = {2000}
}

@article{temlyakov2003_nonlinearapprox,
  title   = {Nonlinear Methods of Approximation},
  author  = {Temlyakov, Vladimir N.},
  journal = {Foundations of Computational Mathematics},
  volume  = {3},
  number  = {1},
  pages   = {33--107},
  year    = {2003}
}

@inproceedings{norrenbrock2025_chiqpm,
  title     = {{CHiQPM}: Calibrated Hierarchical Interpretable Image Classification},
  author    = {Norrenbrock, Thomas and Kaiser, Timo and Biswas, Sovan and Kose, Neslihan and Manuvinakurike, Ramesh and Rosenhahn, Bodo},
  booktitle = {Advances in Neural Information Processing Systems (NeurIPS)},
  year      = {2025}
}


\clearpage
\appendix
\section{Appendix}
\label{appendix}

\subsection{Methodology Details}
\label{appendix:methodology}
This section provides expanded derivations and implementation details that supplement the main paper

\subsubsection{Parametric Dictionary Library}
\label{appendix:parametric_dictionary_library}
 
\paragraph{ExponentialAtom: Numerical Stability via Reparameterization.}
A naive parameterization $A e^{rt}$ causes severe numerical problems during L-BFGS optimization: for large $|r|$ and a long horizon $T$, the output explodes, leading to an ill-conditioned Hessian and diverging gradient steps. 
The conditional time-shift in the main paper resolves this by ensuring the exponent is always non-positive:
\begin{equation}
    \max_{t \in [0, T]} \exp(\cdots) = e^0 = 1
\end{equation}
This bounds the unscaled exponential to $[0, 1]$ and completely decouples $A$ from $r$. 
The amplitude $A$ then directly represents the peak value, equal to the signal at $t = T$ for growth and at $t = 0$ for decay, which is both physically interpretable and well-conditioned for optimization~\cite{boyd2004_convexoptim}.
 
\paragraph{StepAtom: Scale-Invariant Sharpness.}
The normalization factor $\frac{8}{T}$ in the StepAtom makes $\kappa$ scale-invariant across time series of different lengths. 
Since the sigmoid completes the majority of its transition between inputs $-4$ and $+4$, multiplying by $\frac{8}{T}$ ensures that at $\kappa = 1$ the transition spans the full observation window. 
Increasing $\kappa$ sharpens the step independently of whether the series spans milliseconds or years, so the optimizer behaves consistently across datasets.
 
\paragraph{SquareAtom: Duty Cycle Threshold Mechanism.}
The term $\cos(\pi D)$ in the SquareAtom acts as a learnable threshold on the underlying sinusoid before the $\tanh$ activation. 
Increasing $D$ toward $1$ lowers the threshold, widening the active pulse phase, while decreasing $D$ toward $0$ narrows it. 
As $\kappa \to \infty$ the atom converges exactly to $A \cdot \text{sign}(\sin(2\pi f t + \varphi) - \cos(\pi D))$, recovering an ideal square wave with the specified duty cycle~\cite{yu1997_pwm}.
 
\paragraph{SawtoothAtom: Phase-Shifted Construction.}
The two arguments $\varphi_1 = \frac{1}{4}(2\pi f t + \varphi - \pi)$ and $\varphi_2 = \frac{1}{4}(2\pi f t + \varphi + \pi)$ represent the same fundamental frequency sampled at two opposing phase offsets. 
Each contributes one half of the sawtooth profile through the $\tanh$-weighted sinusoidal product, and their sum produces the asymmetric rise-fall pattern~\cite{noble1962_sawtoothinbio}. 
Like the other discontinuous atoms, the sharpness $\kappa$ controls the transition steepness during optimization.
 
\subsubsection{Dictionary Completeness}
\label{appendix:completeness}

We formalize the claim from \autoref{sec:parametric_dictionary_library} that any discrete signal in $\mathbb{R}^N$ is exactly representable as a finite sum of SineAtoms, and therefore as a finite sum of CDP atoms.

\paragraph{Setup.}
Let $\mathbf{t} = (t_0, t_1, \dots, t_{N-1})^\top$ denote the discrete sampling grid with $t_n = n \Delta t$ for some sampling interval $\Delta t > 0$. The SineAtom family, written componentwise on the grid, is
\begin{equation}
    \psi_{\text{sine}}(\mathbf{t}; A, f, \varphi, b) = A \sin(2\pi f \mathbf{t} + \varphi) + b
\end{equation}
where $A, f, \varphi, b \in \mathbb{R}$. The Type-II Discrete Cosine Transform (DCT-II) basis on $\mathbb{R}^N$ consists of the vectors
\begin{equation}
    \mathbf{c}_k[n] = \cos\!\left(\frac{\pi (2n+1) k}{2N}\right), \qquad n = 0, 1, \dots, N-1, \qquad k = 0, 1, \dots, N-1
\end{equation}
The set $\{\mathbf{c}_k\}_{k=0}^{N-1}$ is orthogonal under the standard inner product on $\mathbb{R}^N$, has full rank $N$, and therefore spans $\mathbb{R}^N$~\cite{rao2014_dct}.

\paragraph{Theorem (Completeness).}
For any signal $\mathbf{y} \in \mathbb{R}^N$, there exist coefficients $\{\alpha_k\}_{k=0}^{N-1}$ and parameter tuples $\{(A_k, f_k, \varphi_k, b_k)\}_{k=0}^{N-1}$ such that
\begin{equation}
    \mathbf{y} = \sum_{k=0}^{N-1} \alpha_k \, \psi_{\text{sine}}(\mathbf{t}; A_k, f_k, \varphi_k, b_k)
\end{equation}
exactly on the sampling grid $\mathbf{t}$.

\paragraph{Proof.}
The proof proceeds in two steps.

\textit{Step 1. Each DCT-II basis vector is a SineAtom evaluated on the grid.}
Using the identity $\cos(x) = \sin(x + \pi/2)$, we can write
\begin{equation}
    \mathbf{c}_k[n] = \cos\!\left(\frac{\pi (2n+1) k}{2N}\right) = \sin\!\left(\frac{\pi (2n+1) k}{2N} + \frac{\pi}{2}\right)
\end{equation}
Substituting $t_n = n \Delta t$ and matching to the SineAtom argument $2\pi f t_n + \varphi$ yields the parameter assignment
\begin{equation}
    A_k = 1, \qquad f_k = \frac{k}{2 N \Delta t}, \qquad \varphi_k = \frac{\pi k}{2N} + \frac{\pi}{2}, \qquad b_k = 0
\end{equation}
which gives $\psi_{\text{sine}}(\mathbf{t}; A_k, f_k, \varphi_k, b_k) = \mathbf{c}_k$ exactly on the sampling grid for every $k \in \{0, 1, \dots, N-1\}$. Each frequency $f_k$ lies in $\left[0, \frac{N-1}{2N\Delta t}\right]$, which is below the Nyquist frequency $\frac{1}{2\Delta t}$ and therefore within the admissible parameter range of the SineAtom, where the constant $k = 0$ basis vector is realized in CDP by the bias term rather than a zero-frequency sinusoid.

\textit{Step 2. The DCT-II basis spans $\mathbb{R}^N$.}
Since $\{\mathbf{c}_k\}_{k=0}^{N-1}$ is an orthogonal full-rank basis of $\mathbb{R}^N$, every signal $\mathbf{y} \in \mathbb{R}^N$ admits an exact expansion
\begin{equation}
    \mathbf{y} = \sum_{k=0}^{N-1} \alpha_k \mathbf{c}_k
\end{equation}
where $\alpha_k = \langle \mathbf{y}, \mathbf{c}_k \rangle / \|\mathbf{c}_k\|_2^2$ are the standard DCT coefficients.

\textit{Combining the two steps.}
Substituting the SineAtom representation of $\mathbf{c}_k$ from Step 1 into the DCT expansion from Step 2 yields the claimed sum of $N$ SineAtoms with continuous parameters.

\paragraph{Remarks.}
The proof uses only the SineAtom family, so completeness holds even before the trend, transient, and waveform atom families are added. The remaining atoms enrich the dictionary in the sense that signals with non-sinusoidal physical structure can be represented by far fewer than $N$ atoms. The completeness statement is therefore about representability, not about sparsity. Sparsity in the CDP framework is an empirical property that depends on the match between the atom library and the signal generating process, and is evaluated in \autoref{sec:reconstruction_eval}.
Beyond representability, completeness supplies the hypothesis of the convergence guarantee established next: it ensures that the atomic norm $|\mathbf{y}|_{\mathcal{A}_1(\mathcal{D})}$ is finite for every $\mathbf{y} \in \mathbb{R}^N$.

\subsubsection{From Completeness to Conditional Convergence}
\label{appendix:woga}

Completeness states that the dictionary is rich enough to represent any signal.
It does not state that the greedy procedure actually approaches such a representation.
This section closes that gap.
We show that CDP follows the template of the weak orthogonal greedy algorithm (WOGA)~\cite{temlyakov2000_weakgreedy,temlyakov2003_nonlinearapprox}, a classical scheme for greedy approximation in which each step is allowed to select a good atom rather than the best one.
The value of this connection is a residual decay bound whose only unknown is one number per iteration, and this number can be computed exactly from every fitted model.
The non-convexity of the atom search therefore does not invalidate the analysis, it only determines how large these measured numbers turn out to be.

\paragraph{Setup.}
Let $\mathcal{D} = \{ \psi(\cdot;\theta)/\|\psi(\cdot;\theta)\|_2 : a \in \mathcal{A},\, \theta \in \Theta_a \}$ denote the continuous dictionary with every atom rescaled to unit norm, the standard normalization in greedy approximation theory.
Each parameter domain $\Theta_a$ can be taken compact: frequencies lie in a closed interval below the Nyquist limit enforced by the barrier penalties, temporal anchors lie in the observation window, and amplitudes and biases can be restricted to a bounded box without loss, since the least-squares optimal coefficient of any unit-norm atom is bounded by $\|\mathbf{y}\|_2$ through the Cauchy-Schwarz inequality.
The map $\theta \mapsto \psi(\cdot;\theta)$ is continuous in $L_2$, so suprema over $\mathcal{D}$ are attained.
Let $\mathbf{r}_k$ denote the residual after $k$ greedy iterations ($\mathbf{r}_0 = \mathbf{y}$), and let $\boldsymbol{\varphi}_k \in \mathcal{D}$ be the unit-norm render of the atom accepted at iteration $k$.
The quantity we track is the realized per-step ratio
\begin{equation}
    t_k' \;=\; \frac{|\langle \mathbf{r}_{k-1}, \boldsymbol{\varphi}_k \rangle|}{\|\mathbf{r}_{k-1}\|_2^2}.
    \label{eq:realized_ratio}
\end{equation}
which measures how strongly the accepted atom correlates with the residual it was fitted to.

\paragraph{Lemma (per-step decrease).}
The first step of the analysis is to show that every accepted atom is guaranteed to reduce the residual by a quantifiable amount.
After acceptance of atom $k$, the residual satisfies
\begin{equation}
    \|\mathbf{r}_k\|_2^2 \;\leq\; \|\mathbf{r}_{k-1}\|_2^2 - \langle \mathbf{r}_{k-1}, \boldsymbol{\varphi}_k \rangle^2.
    \label{eq:perstep_decrease}
\end{equation}
\textit{Proof.}
The right-hand side of \eqref{eq:perstep_decrease} is exactly the error left after the best least-squares fit of the single direction $\boldsymbol{\varphi}_k$ to the residual, so the claim is that every accepted atom does at least as well as this fit.
This holds for a simple reason.
Amplitude and bias enter every atom render linearly, so setting the gradient of the fitting objective to zero in these two coordinates is the same as solving their least-squares normal equations.
An accepted atom whose amplitude and bias are first-order stationary is therefore the exact least-squares fit of its own shape to the residual, which achieves the decrease \eqref{eq:perstep_decrease} with all earlier atoms unchanged.
CDP drives these coordinates to stationarity with L-BFGS under strong Wolfe line search, and the condition can also be enforced exactly by one closed-form rescaling of the accepted atom.
All subsequent refinement stages, namely the full refinement of the accepted atom and the interleaved global refinement of all atoms, accept only steps that decrease the objective, so they can only shrink the residual further. \hfill$\square$

\paragraph{Theorem (computable residual bound).}
Accumulating the guaranteed decreases over all iterations yields a bound on the final residual.
Under \eqref{eq:perstep_decrease}, the residual after $K$ iterations satisfies
\begin{equation}
    \frac{1}{\|\mathbf{r}_K\|_2^2} \;\geq\; \frac{1}{\|\mathbf{y}\|_2^2} + \sum_{k=1}^{K} (t_k')^2.
    \label{eq:computable_bound}
\end{equation}
\textit{Proof.}
Writing $a_k = \|\mathbf{r}_k\|_2^2$ and substituting \eqref{eq:realized_ratio} into \eqref{eq:perstep_decrease} gives $a_k \leq a_{k-1}\bigl(1 - (t_k')^2 a_{k-1}\bigr)$, where $(t_k')^2 a_{k-1} = \langle \mathbf{r}_{k-1}, \boldsymbol{\varphi}_k\rangle^2 / a_{k-1} \leq 1$ by Cauchy-Schwarz.
Taking reciprocals and using the elementary inequality $\frac{1}{x(1-\varepsilon x)} \geq \frac{1}{x} + \varepsilon$ for $\varepsilon x < 1$ yields $a_k^{-1} \geq a_{k-1}^{-1} + (t_k')^2$.
Summing over $k$ gives \eqref{eq:computable_bound}. \hfill$\square$

The point of this bound is that it contains no unknown constants.
Only the signal energy and inner products of computed vectors appear, so the bound can be evaluated exactly on every fitted model, and no supremum over the continuous dictionary is involved.

\paragraph{Corollary (a priori rate).}
The classical WOGA analysis asks how close each greedy step comes to the best atom in the entire dictionary.
Call the scouting step $\tau$-approximate, $\tau \in (0,1]$, if $|\langle \mathbf{r}_{k-1}, \boldsymbol{\varphi}_k\rangle| \geq \tau \sup_{g \in \mathcal{D}} |\langle \mathbf{r}_{k-1}, g\rangle|$ at every iteration, meaning the selected atom captures at least a fraction $\tau$ of the correlation that the best atom in $\mathcal{D}$ would achieve.
With the per-step decrease of the lemma alone, CDP is a $\tau$-weak greedy algorithm, for which the classical rates are weaker~\cite{temlyakov2000_weakgreedy}.
It becomes a weak orthogonal greedy algorithm with weakness parameter $\tau$ under the additional condition that the residual is kept orthogonal to the span of all accepted atoms, which corresponds to a joint least-squares update of all amplitudes and biases and is what the interleaved global refinement approximates.
Under this condition, Temlyakov's rate theorem~\cite{temlyakov2000_weakgreedy,temlyakov2003_nonlinearapprox} gives
\begin{equation}
    \|\mathbf{r}_K\|_2 \;\leq\; |\mathbf{y}|_{\mathcal{A}_1(\mathcal{D})} \, \bigl(1 + \tau^2 K\bigr)^{-1/2},
\end{equation}
where $|\mathbf{y}|_{\mathcal{A}_1(\mathcal{D})}$ is the atomic norm of $\mathbf{y}$ with respect to $\mathcal{D}$, a measure of how cheaply $\mathbf{y}$ can be assembled from dictionary atoms.
The completeness theorem of \autoref{appendix:completeness} guarantees that this norm is finite for every $\mathbf{y} \in \mathbb{R}^N$, so the bound applies to every signal.
In practice the computable form \eqref{eq:computable_bound} is the more useful statement, since $\|\mathbf{y}\|_2 \leq |\mathbf{y}|_{\mathcal{A}_1(\mathcal{D})}$ makes it the tighter of the two.
An explicit upper bound on the atomic norm can further be obtained by basis pursuit over any finite sub-dictionary, which can only overestimate it.

\paragraph{Numerical verification.}
We evaluate the bound \eqref{eq:computable_bound} post hoc on fitted models at $K = 10$ on all 13 real-world datasets, recomputing all projections with exact least-squares solves on the fitted atom renders, independently of the optimization path.
Since every quantity in the bound is computed from the realized greedy trajectory, this is a diagnostic of how tightly the guaranteed per-step decreases account for the achieved training error, not an independent proof of convergence.
Three observations support the analysis.
First, the assumption of the lemma holds in practice: the certified projection cascade matches the model's own training error to within $9\%$ relative MSE on every dataset, with a median gap below $1\%$, and the small remaining differences reflect the finite L-BFGS tolerance and the finite sharpness of the training surrogates for discontinuous atoms.
Second, the bound is tight: it tracks the actual training error with a median ratio of $1.12$ and a range of $1.00$ to $1.67$ across datasets, for example $0.23$ against $0.19$ on PEMS04.
Per greedy step, the selected atom captures a median $7\%$ of the remaining residual energy and up to $74\%$, and $92\%$ of steps capture at least $1\%$, with the rare near-zero steps occurring when the interleaved global refinement leaves two atoms nearly collinear, so that the second one contributes almost no incremental energy in the exact cascade.
Third, the scouting step is close to optimal: comparing each selected atom against a randomized reference search over the continuous dictionary with $100$ refined candidates per atom family, the realized weakness parameter $\tau$ has a median of $1.0$ and exceeds $0.8$ in $85\%$ of steps, occasionally beating the reference search itself.
The realized ratios $t_k'$ are also logged during fitting.

\paragraph{Scope.}
The guarantee is conditional: we do not prove global optimality of the non-convex scouting step, and no method in this class does.
Formal sparse recovery and identifiability results for mixed parametric families remain open.
Existing guarantees from atomic norm minimization~\cite{bhaskar2013_atomicnorm,tang2013_compressed,candes2014_superres} cover the sinusoidal case, where the Toeplitz structure applies, and no analogue is known for dictionaries mixing sinusoids, Gaussians, polynomials, and discontinuous waveforms.

\subsubsection{Consistency of the Sharpness Annealing}
\label{appendix:gamma}

The differentiable surrogates for discontinuous atoms are consistent approximations of their ideal sharp-limit renders, in the following sense.

\paragraph{Lemma ($\Gamma$-convergence of the annealed objectives).}
For the square, sawtooth, and step families, let $g_\kappa(\cdot;\theta)$ denote the annealed surrogate with sharpness $\kappa$ and $g_\infty(\cdot;\theta)$ the ideal render.
Assume that the duty cycle of the square family is bounded away from $\{0, 1\}$ and that the transition of the step family lies inside the observation window.
Then:
\begin{enumerate}
    \item[(i)] $\|g_\kappa(\cdot;\theta) - g_\infty(\cdot;\theta)\|_{L_2} \leq C(\theta)\, \kappa^{-1/2}$, where $C(\theta)$ is explicit and diverges only as the duty cycle approaches $0$ or $1$.
    \item[(ii)] The annealed least-squares objectives $F_\kappa$ $\Gamma$-converge to the ideal objective $F_\infty$ on the compact parameter domain. Consequently, every accumulation point of minimizers of $F_\kappa$ as $\kappa \to \infty$ is a minimizer of $F_\infty$, and the minimal values converge.
\end{enumerate}
\textit{Proof.}
(i) $\tanh(\kappa u)$ differs appreciably from $\operatorname{sign}(u)$ only where $|u| = O(1/\kappa)$.
The argument $u(t) = \sin(2\pi f t + \varphi) - \cos(\pi D)$ crosses zero with derivative bounded below precisely when $D$ is bounded away from $0$ and $1$, so the set where the surrogate and the ideal render differ has measure $O(1/\kappa)$ per period.
On this set the difference is bounded by $2A$, which gives an $L_2$ error of $O(\kappa^{-1/2})$.
The constant degrades as the crossing derivative vanishes, i.e., at extreme duty cycles.
(ii) Since the objective is quadratic in the render, (i) implies uniform convergence of $F_\kappa$ to $F_\infty$ on the compact domain, and uniform convergence on a compact set implies $\Gamma$-convergence together with convergence of minimizers. \hfill$\square$

\paragraph{Empirical confirmation.}
The finite-sharpness gap predicted by (i) is measured at approximately $1\%$ at the final sharpness $\kappa = 50$ for moderate duty cycles, and grows at extreme duty cycles, exactly the regime the constant $C(\theta)$ identifies.
The theory thus predicts the one regime where the annealed approximation is visibly loose.
 
\subsubsection{Data-Driven Proposal Distribution}
 \label{appendix:data_driven_priors}
The main paper states that localization parameters are bounded within the active time window and amplitudes are scaled to the residual energy. 
The motivation is identifiability: a GaussianAtom with center $\mu$ placed far outside $[0, T]$ contributes near-zero variance within the observation window and is functionally indistinguishable from a constant, producing zero-gradient regions that stall L-BFGS. 
Strictly bounding $\mu$ and $t_{\text{step}}$ within $[0, T]$ at initialization prevents this degenerate case.
 
\subsubsection{Soft Barrier Penalty Details}
 \label{appendix:limit_losses}
To enforce structural validity, the MSE objective is augmented with a soft barrier penalty term:
\begin{equation}
    \mathcal{L}(\theta_k) = \operatorname{MSE}(\theta_k) + w_{\text{limit}} \sum_{p \in \theta_k}
    \mathcal{L}_{\text{limit}}(p)
\end{equation}
where $p \in \theta_k$ denotes any scalar entry of the parameter tuple $\theta_k$, and $w_{\text{limit}} > 0$ controls the barrier strength.
The full set of per-atom constraints enforced by $\mathcal{L}_{\text{limit}}$ is:
\begin{itemize}
    \item \textbf{Frequency} $f$: penalized as $f \to 0$ (DC) or $f > f_{\text{Nyquist}}$ (aliasing).
    \item \textbf{Gaussian width} $\sigma$: penalized as $\sigma \to 0$ to suppress impulse-like artifacts.
    \item \textbf{Temporal anchors} $\mu$, $t_{\text{step}}$: penalized outside $[0, T]$ for the identifiability reason above.
    \item \textbf{Duty cycle} $D$: bounded to $(0, 1)$ to keep the SquareAtom physically valid.
\end{itemize}
The L-BFGS $i$-th step descent direction $p_i = -H_i \nabla \mathcal{L}(\theta_i)$ uses an approximate inverse Hessian $H_i$ built from a rolling window of gradient differences. 
This second-order information implicitly rescales each parameter dimension by its local curvature, which is critical for the CDP objective where frequency and amplitude sensitivities can differ by several orders of magnitude.


\subsection{Synthetic Parameter Recovery}
\label{appendix:synthetic_eval}

We provide the exact generative equations and recovered parameters for all six synthetic experiments from \autoref{sec:synthetic_eval}.
Both input and target signals are corrupted with additive Gaussian noise, $\tilde{y}(t) = y(t) + \varepsilon$, $\varepsilon \sim \mathcal{N}(0, \sigma^2)$ with $\sigma = 0.1$, and all MSE values are computed against the clean ground truth $y(t)$.
This setup jointly tests structural decomposition and denoising, providing a stronger evaluation than noise-free reconstruction alone.

\paragraph{Linear Trend:}
\begin{align}
    y &= 0.5\cdot t \\
    \hat{y} &= 0.5\cdot t - 0.0008
\end{align}
MSE: $6.21\mathrm{e}{-7}$. 
The LinearAtom recovers both slope and bias accurately, confirming that greedy selection correctly identifies and isolates trend parameters even under noise.

\paragraph{Sine Wave:}
\begin{align}
    y &= 1.5 \cdot \sin(2\pi \cdot 4.0 \cdot t) \\
    \hat{y} &= 1.4996 \cdot \sin(2\pi \cdot 4.0006 \cdot t - 0.0059) - 0.005
\end{align}
MSE: $2.45\mathrm{e}{-5}$. 
All three parameters, namely amplitude, frequency, and phase, are recovered to high precision, validating the FFT-based spectral proposal mechanism.

\paragraph{Differentiable Square Wave:}
\begin{align}
    y &= 1.5 \cdot \text{sign}\left(\sin(2\pi \cdot 4.0 \cdot t) - \cos(0.7\pi)\right) \\
    \hat{y} &= 1.514 \cdot \tanh\!\left(\frac{\kappa}{2} \cdot (\sin(2\pi \cdot 3.996 \cdot t + \varphi) - \cos(0.7001\pi))\right)
\end{align}
MSE: $2.40\mathrm{e}{-4}$.
The duty cycle is recovered at $D = 0.7001$, within $\sim 0.01\%$ of the true value, demonstrating that spectral annealing successfully sharpens the waveform edges without trapping the optimizer in early local minima.

\paragraph{Off-Grid Frequency Superposition:}
\begin{align}
    y &= 1.5 \cdot \sin(2\pi \cdot 5.8 \cdot t) + 1.0 \cdot \sin(2\pi \cdot 10.3 \cdot t + 0.5) \\
    \hat{y} &= 1.5 \cdot \sin(2\pi \cdot 5.8001 \cdot t) + 1.0 \cdot \sin(2\pi \cdot 10.3001 \cdot t + 0.5)
\end{align}
MSE: $7.00\mathrm{e}{-5}$. 
Because both frequencies fall between FFT bins, standard OMP produces a reconstruction MSE of $0.36$ due to severe spectral leakage. 
CDP resolves both components to within $0.0002\%$ through continuous frequency optimization.

\paragraph{Damped Sine Wave:}
\begin{align}
    y &= 2.0 \cdot \sin(2\pi \cdot 5 \cdot t) \cdot \exp(-1.5 \cdot t) \\
    \hat{y} &= 2.04 \cdot \sin(2\pi \cdot 5.0049 \cdot t + 0.0199) \cdot \exp(-1.3902 \cdot t) - 0.07
\end{align}
MSE: $6.32\mathrm{e}{-4}$. 
The correct carrier-modulator pairing emerges from the greedy selection without supervision. 
OMP, whose Fourier dictionary cannot represent the multiplicative carrier-envelope coupling, yields a reconstruction MSE of $0.07$ and a forecasting MSE of $0.5$.

\paragraph{Non-Stationary Amplitude Shift:}
Here $H(t)$ denotes the Heaviside step function and $\sigma(t)$ the sigmoid. 
\begin{align}
    y &= \sin(2\pi \cdot 4 \cdot t) \cdot 1.3 \cdot H(t - 0.5) + 0.2 \\
    \hat{y} &= 1.2875 \cdot \sin(2\pi \cdot 4.0574 \cdot t - 0.15) \cdot \sigma\!\left(\tfrac{8 \cdot 50}{0.8}(t - 0.505)\right) + 0.19
\end{align}
MSE: $8.85\mathrm{e}{-3}$. 
The $0.8$ in the sigmoid's steepness term $\frac{8 \cdot 50}{0.8}$ equals $T_{\text{train}} = 0.8$, the length of the training interval under an 80/20 train-test-split of the full unit interval $T = 1.0$.
CDP reconstructs the regime change with a single ModulatedAtom, placing the transition at $t = 0.505$, within $1\%$ of the true shift point. 
OMP's static dictionary is ill-suited for non-stationary shifts, resulting in a reconstruction MSE of $0.31$ and a forecasting MSE of $1.38$.


\subsection{Robustness and Stability}
\label{appendix:robustness}

This section collects the analyses of CDP's sensitivity to initialization, hyperparameters, noise, and library design.

\paragraph{Stability across random seeds:}
At an atom budget of $K=5$ across all 13 real-world datasets and 10 seeds, we rank-match the recovered periodic atoms across seeds and measure the relative standard deviation of their frequencies: $76\%$ of matched atoms vary by less than $1\%$, and $81\%$ by less than $5\%$.
The selected atom-type multiset is identical across all seeds on 8 of 13 datasets.
The recovered decompositions are therefore substantially a property of the signal rather than of the random initialization.

\paragraph{Stability across time windows:}
To verify that recovered atoms reflect persistent signal structure rather than idiosyncrasies of a particular observation window, we refit CDP at $K=5$ on ten overlapping sub-windows per dataset.
Each sub-window spans $80\%$ of the training portion, the start offsets are evenly spaced over the remaining $20\%$, and the fit seed is held fixed so that the observation window is the only varying factor.
Periodic atoms are matched across windows by clustering their frequencies with a $2\%$ relative tolerance, and a component is called persistent if it appears in at least eight of the ten windows.
Transient atoms (Gaussian, step) are localized by construction and are excluded from the matching.
Across the 13 datasets this yields 25 persistent periodic components, and their frequencies are essentially window-invariant: $96\%$ vary by less than $1\%$ relative standard deviation across windows (median $0.02\%$), and all vary by less than $5\%$.
The waveform-type label at a persistent frequency is less stable and matches the modal type in $76\%$ of appearances, since neighboring waveform families, for example square and sawtooth atoms at the same fundamental frequency, can approximate the same component.
On Electricity and Weather no component reaches the persistence threshold at this atom budget, as their energy is distributed over many components of comparable strength that trade places across windows.

\paragraph{Noise robustness:}
On a more complex synthetic dataset with known ground truth, we evaluate recovery under increasing noise levels.
CDP recovers the main components for SNR values down to $2$.
Below that, the atoms with lower power degrade gradually rather than failing abruptly, which indicates graceful behavior at the noise floor.

\paragraph{Hyperparameter sensitivity:}
We ran additional configurations on all 13 real datasets varying the key hyperparameters jointly: scouting batch size increased $5\times$ (10 to 50), scouting epochs $10\times$ (5 to 50), and global refinements per fit from 2 up to one refinement per atom.
The resulting median training reconstruction differs by less than $3\%$ at matched atom budgets, indicating robustness to these choices.

\paragraph{Atom-library ablation:}
The effect of restricting CDP to a sinusoid-only library is analyzed as part of the morphology breakdown in \autoref{appendix:baselines}, which separates library coverage from optimizer quality per dataset.

\paragraph{Out-of-library structure:}
CDP removes discretization error only for structure expressible in its library.
When a signal contains structure outside every atom family, such as a continuously varying instantaneous frequency, the fit degrades toward a best-in-library approximation composed of locally fitted atoms, and the extracted parameters describe that approximation rather than the generating process (cf.\ \autoref{sec:limitations}).


\subsection{Atom-Based Curriculum Learning}
\label{appendix:curriculum_learning}

This section provides a detailed description of the curriculum learning procedure introduced in \autoref{sec:curriculum_eval}, including the evaluation protocol, the target construction, the curriculum weighting schedule, and the S4D sequence model used as the downstream learner.
 
\subsubsection{Overview}
 
The core idea is to use the $K$ atoms fitted by CDP as a structured sequence of learning targets for a neural sequence model. 
Rather than training directly on the raw signal, which presents the model with the full complexity of all components simultaneously, the curriculum introduces atoms one at a time. 
Each atom is smoothly blended into the training target via a sigmoid-shaped weight that transitions from zero to one over a configurable number of iterations.
In the final phase, the model is fine-tuned on the normalized real signal.

\subsubsection{Evaluation Protocol}
\label{appendix:cl_protocol}

The evaluation protocol, the compared curriculum targets, and the aggregation statistics are described in \autoref{sec:curriculum_eval}.
For NOMP and SFW the stored constant offset enters as an additional first component of the staged decomposition.
The confidence intervals of the curriculum and standard training conditions overlap on most individual datasets, so the per-dataset effect is modest but consistent in direction across the benchmark.

\subsubsection{Curriculum Target Construction}
\label{appendix:cl_target}
 
Given a decomposition into $K$ components $\{\psi_k\}_{k=1}^K$, the curriculum target at iteration $i$ is defined as the weighted sum:
\begin{equation}
    y^{\text{curr}}(\mathbf{t}; i) = \sum_{k=1}^{K} w_k(i) \cdot \psi_k(\mathbf{t}; \theta_k)
\end{equation}
where $w_k(i) \in [0, 1]$ is the curriculum weight of component $k$ at iteration $i$, computed from the current iteration index as described below.
In practice each component enters as its stored per-component reconstruction from the fitted decomposition, which lets the non-parametric baseline components be staged through the same formula.
 
\subsubsection{Curriculum Weighting Schedule}
 \label{appendix:cl_weight_schedule}
The total training budget is divided into $T_{\text{curr}} = T \cdot (1 - \rho)$ curriculum iterations and $T_{\text{real}} = T \cdot \rho$ fine-tuning iterations on the real signal, where $T$ is the total number of iterations and $\rho$ is the real-data ratio hyperparameter.
The curriculum iterations are divided equally among $K$ atoms, giving $T_k = T_{\text{curr}} / K$ iterations per atom.
The activation shift for atom $k$ is $s_k = k \cdot T_k$ for $k = 0, \dots, K-1$, so the first atom is active from the start of training.
 
The weight of atom $k$ at iteration $i$ is computed as:
\begin{equation}
    \tilde{w}_k(i) = \sigma\!\left(\frac{6(i - s_k)}{\alpha \cdot T_k} + 6\right)
\end{equation}
\begin{equation}
    w_k(i) = \begin{cases} 1 & \text{if } \tilde{w}_k(i) > \sigma(\alpha \cdot T_k) \\ 0 & \text{if } \tilde{w}_k(i) < \sigma(-\alpha \cdot T_k) \\ \tilde{w}_k(i) & \text{otherwise} \end{cases}
\end{equation}
where $\sigma$ is the logistic sigmoid and $\alpha$ is the \texttt{curriculum\_smoothness} hyperparameter. 
Weights above $\sigma(\alpha \cdot T_k)$ or below $\sigma(-\alpha \cdot T_k)$ are clamped to exactly $1$ or $0$. At the settings used, these thresholds are numerically indistinguishable from $1$ and $0$, so the clamping acts as a numerical saturation guard rather than an active mechanism.

A small $\alpha$ produces a sharp step-like transition, while a larger $\alpha$ yields a gradual blend.

\subsubsection{Training Phases}
 \label{appendix:cl_training}
Training proceeds in two phases. During the curriculum phase (iterations $0$ to $T_{\text{curr}}$), the model is trained on $y^{\text{curr}}(\mathbf{t}; i)$ computed from the weighted atom sum.
During the fine-tuning phase (iterations $T_{\text{curr}}$ to $T$), the target is replaced by the z-score normalized ground truth signal $\tilde{y}(t) = (y(t) - \mu_{\text{train}}) / \sigma_{\text{train}}$, allowing the model to adapt to residual structure not captured by the CDP decomposition. 
The loss in both phases is the mean squared error between the model output and the current target.
The model maps the normalized time index to the signal value, so out-of-sample predictions are obtained by evaluating the trained model on the continuation of the time axis, and the reported test MSE is computed on the segment beyond the training portion.
 
\subsubsection{The S4D Sequence Model}
 \label{appendix:cl_s4d_model}
The downstream learner is a vanilla S4D model, a stack of $L$ S4D blocks~\cite{gu2022_s4d} with a pointwise input projection and a linear output projection. 
The model maps a univariate time series $\mathbf{x} \in \mathbb{R}^{N \times 1}$ to a reconstruction $\hat{\mathbf{x}} \in \mathbb{R}^{N \times 1}$.
 
Each S4D block contains a structured state space kernel, a local depthwise convolution, a GELU activation followed by a gated linear unit output projection, and a residual connection with layer normalization.
The S4D kernel parameterizes a diagonal linear time-invariant system. 
For a block with model dimension $H$ and state dimension $N$, the kernel at sequence length $\ell \in \{0,1,...,L-1\}$ is computed as:
\begin{equation}
    k_h[\ell] = 2\,\text{Re}\!\left(\sum_{n=1}^{N/2} \tilde{C}_{hn} \cdot e^{A_{hn}\,\ell}\right), \qquad \tilde{C}_{hn} = C_{hn} \cdot \frac{e^{A_{hn} \Delta_h} - 1}{A_{hn}}
\end{equation}
where $C_{hn} \in \mathbb{C}$ are the output projection weights and $\Delta_h \in \mathbb{R}_{>0}$ is the per-channel time step. 
To enforce stability and positivity through unconstrained optimisation, both the time step and the magnitude of the real part of the state matrix are parameterised in log-space,
\begin{equation}
    A_{hn} = -\exp(\log A^{\text{re}}_{hn}) + i\pi n, \qquad \Delta_h = \exp(\log \Delta_h)
\end{equation}
so that the learned variables $\log A^{\text{re}}_{hn}$ and $\log \Delta_h$ are real-valued and free of constraints. The imaginary part $i\pi n$ of $A_{hn}$ assigns each state a fixed oscillatory frequency, allowing the kernel to represent periodic patterns efficiently. 
The real part is held strictly negative by construction, which guarantees that the underlying continuous-time system is stable and that the kernel decays over $\ell$.

The kernel is applied to the input via FFT convolution:
\begin{equation}
    \mathbf{y}_h = \text{iFFT}\!\left(\text{FFT}(\mathbf{x}_h) \odot \text{FFT}(\mathbf{k}_h)\right) + \mathbf{x}_h \cdot D_h
\end{equation}
where $D_h$ is a learned skip connection weight. 
A local depthwise convolution with kernel size 3 is applied after the global convolution to help capture sharp local transitions such as those produced by SquareAtom and StepAtom. 
The output then passes through a gated linear unit and a residual connection.
 
The full model architecture and the hyperparameters for the curriculum learning are summarized in \autoref{tab:s4d_arch} and \autoref{tab:curriculum_hyperparams} respectively.
 
\begin{table}[htbp]
\centering
\caption{\textbf{AtomS4DModel architecture.} Default hyperparameters used in all curriculum learning experiments.}
\label{tab:s4d_arch}
\begin{tabular}{lll}
\toprule
\textbf{Component} & \textbf{Details} & \textbf{Default} \\ \midrule
Input projection & Conv1d($1 \to d_{\text{model}}$, kernel 1) & $d_{\text{model}} = 128$ \\
S4D blocks & $n_{\text{layers}}$ stacked S4DBlocks & $n_{\text{layers}} = 4$ \\
State dimension & Per-block S4D state size & $d_{\text{state}} = 64$ \\
Local conv & Depthwise Conv1d(kernel 3, padding 1) & per block \\
Output gate & GLU via Conv1d($d_{\text{model}} \to 2d_{\text{model}}$) & per block \\
Dropout & Applied after GLU & $p = 0.1$ \\
Output projection & Linear($d_{\text{model}} \to 1$) & \\
Time step range & log-spread for different time resolutions & $\Delta \in [10^{-3}, 10^{-1}]$ \\
Optimizer & Adam & lr $= 10^{-3}$ \\ \bottomrule
\end{tabular}
\end{table}
 
\begin{table}[htbp]
\centering
\caption{\textbf{Curriculum learning hyperparameters} used in all experiments.}
\label{tab:curriculum_hyperparams}
\begin{tabular}{lll}
\toprule
\textbf{Hyperparameter} & \textbf{Description} & \textbf{Default} \\ \midrule
$T$ & Total training iterations & 1000 \\
$K$ & Total number of atoms & 10 \\
$\rho$ & Fraction of iterations on real data & 0.1 \\
$\alpha$ & Curriculum transition smoothness & 0.2 \\ \bottomrule
\end{tabular}
\end{table}


\subsection{Extrapolation and Forecasting Performance}
\label{appendix:forecasting_eval}

\begin{figure}[htbp]
\begin{center}
\includegraphics[width=1.0\columnwidth, clip]{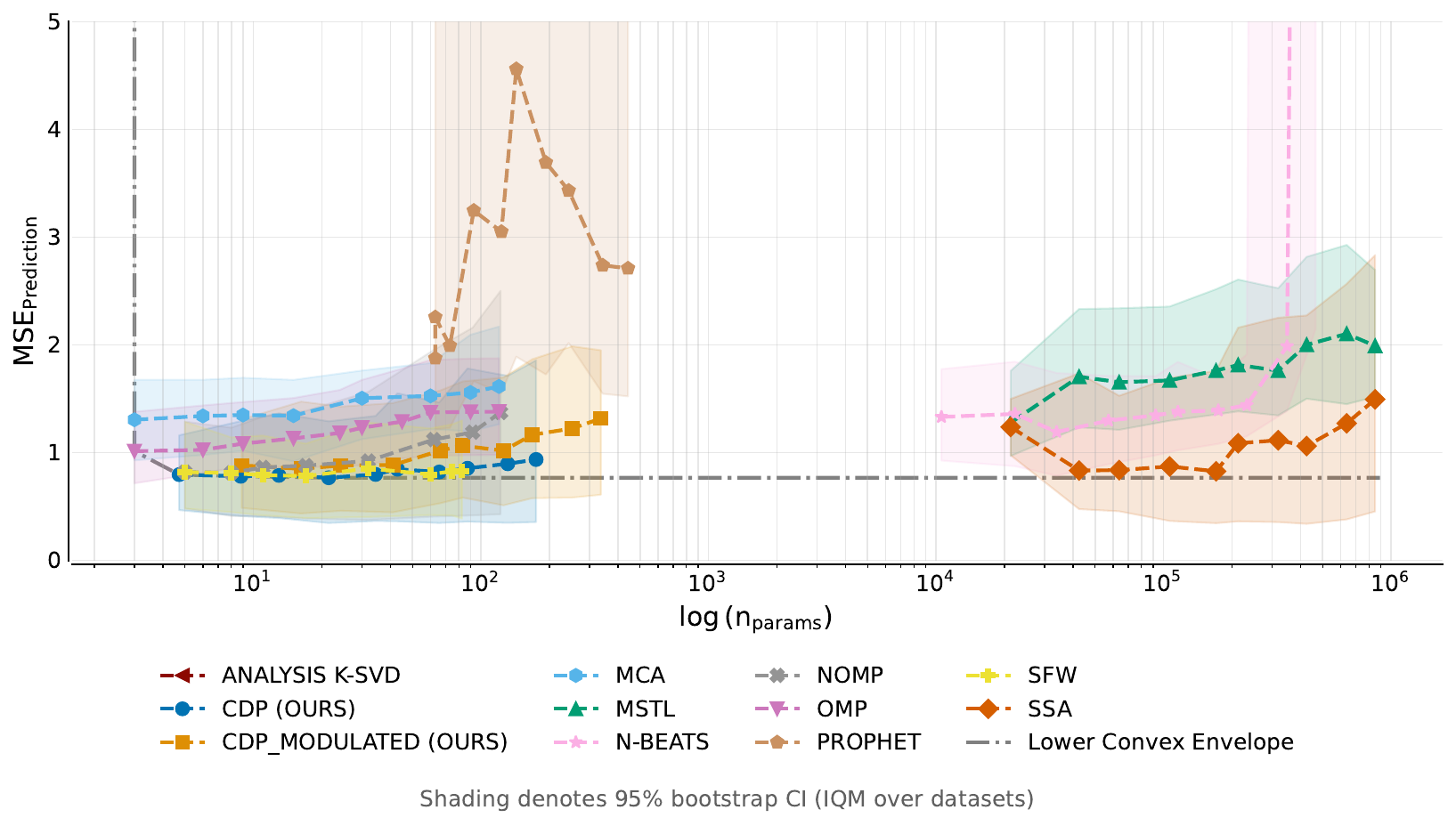}
\caption{\textbf{Pareto frontier of model complexity versus out-of-sample forecasting error.} The x-axis indicates the parameter count on a logarithmic scale. The y-axis shows the out-of-sample forecasting MSE. CDP and CDP (Modulated) maintain their performance at the sparse end of the frontier, achieving competitive forecasting accuracy with far fewer parameters than deep learning baselines. The continuous and structured baselines (NOMP, SFW, MCA, Analysis K-SVD) are included, with Analysis K-SVD lying outside the displayed range (median test error above $28$, see \autoref{appendix:baselines}).}
\label{fig:forecasting_evaluation}
\end{center}
\end{figure}%

Because CDP fits closed-form parametric functions to the training signal, it can extrapolate deterministic periodicities over long horizons without the autoregressive error accumulation typical of deep learning models.
As shown in \autoref{fig:forecasting_evaluation}, CDP and CDP (Modulated) maintain a favorable position on the forecasting Pareto frontier across the LTSF benchmark suite with a stable forecasting MSE of approximately $0.8$–$1.0$.

Prophet is the most unstable baseline, with forecasting MSE peaking at approximately $4.6$ in the $10^2$ parameter range before declining, reflecting its inability to reliably capture non-sinusoidal periodicities.
OMP starts near $1.0$ but drifts upward as more atoms are added, since spectral leakage from off-grid frequencies accumulates rather than cancels across greedy steps.
MSTL and SSA enter only beyond $10^4$ parameters and converge to MSEs of $1.5$–$2.0$, well above CDP.
N-BEATS enters at $10^4$ parameters with high variance and does not consistently improve beyond $1.3$, likely because its non-autoregressive architecture is forced to generate long-range predictions without iterative feedback in this evaluation setting.

Performance degrades when non-stationary domain shifts occur between the train and test sets, since the fitted equations extrapolate the historically identified structure without adapting to unforeseen changes.
This is an inherent property of any model that performs decomposition by fitting analytical structure to historical observations rather than a deficiency specific to CDP.


\subsection{SHAP Analysis}
\label{appendix:shap_eval}

\begin{figure}[htbp]
  \begin{center}
    \includegraphics[width=\columnwidth]{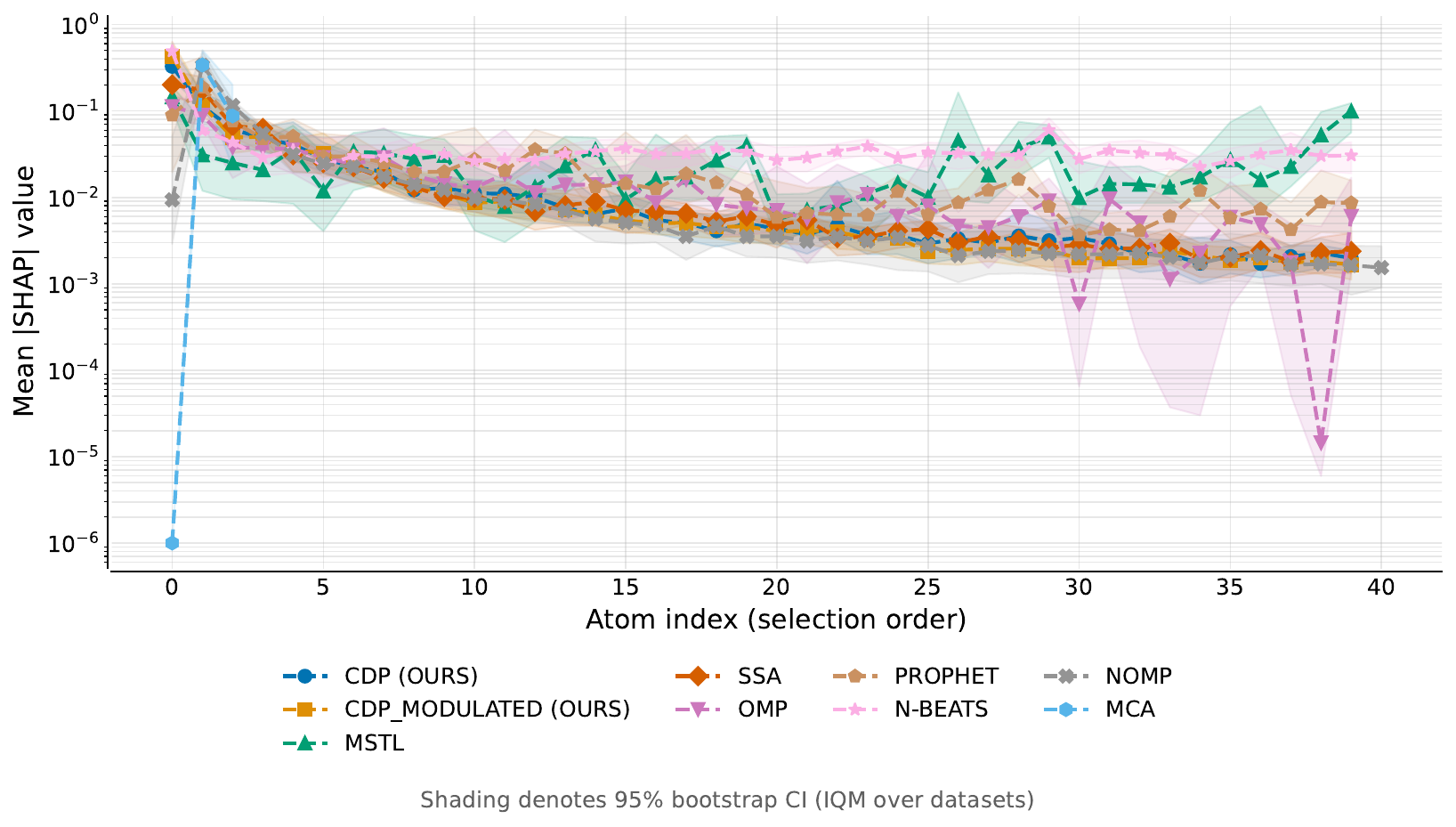}
    \caption{\textbf{Mean absolute SHAP value per atom index} in greedy selection order, averaged over all experiments and datasets (log scale). Shading denotes 95\% bootstrap confidence intervals. SFW is excluded from the figure because its per-atom attributions diverge at larger budgets (see text).}
    \label{fig:shap_per_atom}
  \end{center}
\end{figure}%

We apply SHapley Additive exPlanations (SHAP)~\cite{lundberg2017_shapvalues} to quantify how explanatory weight is distributed across the atoms of each fitted model.
This analysis quantifies the \emph{separation} of components, that is, whether each atom captures a distinct portion of the signal's explainable variance.
It does not verify the \emph{physical identity} of the components, which cannot be established without ground-truth decompositions (cf.\ \autoref{sec:limitations}).
For atom $k$, the Shapley value $\nu_k$ is its average marginal contribution to reconstruction quality over all possible subsets of the active atom set, approximated via Monte Carlo sampling~\cite{vstrumbelj2014_explainingfeatures}:
\begin{equation}
  \nu_k = \sum_{S \subseteq \{1,\dots,K\} \setminus \{k\}} \frac{|S|!(K-|S|-1)!}{K!} \left[ v(S \cup \{k\}) - v(S) \right]
\end{equation}
where $v(S)$ denotes the reduction in MSE achieved by subset $S$ alone.
All SHAP values are computed post hoc from the stored component reconstructions of the final fitted models under this single definition.
Computed this way, each model's values sum exactly to the explained variance of its final fit, which makes the attributions comparable across methods.
\autoref{fig:shap_per_atom} shows the mean absolute SHAP values as a function of the atom index in greedy selection order.

\paragraph{Atom Definition Across Methods:}
Because SHAP requires a fixed set of additive components, we define an "atom" consistently across methods as one column of each model's component reconstruction. 
For CDP and OMP this is unambiguous, since both methods are explicitly additive over their dictionary entries. 
For the structural and learned baselines we adopt the following conventions. 
MSTL contributes one trend atom from the LoESS-smoothed level component and $K-1$ seasonal atoms from periods detected via the FFT of the input signal, matching the standard MSTL output decomposition~\cite{bandara2025_mstl}. 
SSA contributes $K$ rank-one atoms, each obtained by Hankel-averaging a single singular triplet $\sigma_i u_i v_i^\top$ from the trajectory matrix decomposition, sorted by descending singular value. 
N-BEATS contributes one atom per residual block, with the first block corresponding to the trend stack (polynomial basis) and the remaining $K-1$ blocks to the seasonality stack (Fourier basis), following the doubly-residual interpretation introduced by Oreshkin et al.~\cite{oreshkin2019_nbeats}. 
Prophet contributes one trend atom and $K-1$ Fourier seasonality atoms from its native additive model.
NOMP contributes one atom per refined sinusoid, and MCA contributes one component per morphological sub-dictionary (trend, sinusoidal, transient), so MCA has exactly three components at every budget.
For NOMP and MCA the first stored component is the constant offset, which carries no variance after normalization and appears as a near-zero first index.
In all cases the atoms sum to the model's full reconstruction, so SHAP values $\nu_k$ are directly comparable across methods.
Two of the added baselines are excluded.
Analysis K-SVD is not additive over an atom set and admits no comparable attribution.
SFW is additive, but at larger budgets it places near-duplicate atoms with large cancelling weights, so individual atoms carry attributions that diverge by several orders of magnitude and are meaningful only jointly.
Additivity alone therefore does not make components individually interpretable, and we characterize both methods through the Pareto, morphology, and cost analyses in \autoref{appendix:baselines} instead.

\paragraph{Concentration of Explanatory Weight:}
CDP concentrates the most explanatory weight in its leading atoms: the first CDP atom carries roughly three times the contribution of the first OMP atom ($0.33$ against $0.11$ in the IQM over datasets), since a single continuous atom can settle at the exact frequency, amplitude, and phase of a physical component that OMP must spread across neighboring grid atoms.
Along the selection sequence, the continuous methods CDP, CDP (Modulated), and NOMP decay smoothly toward the lowest tail values.
NOMP matches CDP's decay, which is consistent with the morphology analysis in \autoref{appendix:baselines}: clean per-atom concentration is a property of continuous parameter optimization, and CDP's advantage over single-family continuous methods lies in library coverage rather than concentration.

The grid-based and smoothing-based methods behave differently.
OMP exhibits non-monotone contributions along the selection sequence, where an atom selected later contributes more than atoms selected earlier, reflecting the leakage corrections of the grid mismatch problem~\cite{duarte2013_spectralcompressivesensing,pezeshki2015_gridmismatch}.
MSTL shows an irregular profile with upward excursions across the full index range, and N-BEATS and Prophet maintain elevated tails, indicating that their explanatory weight remains distributed rather than isolated in a few components.
SSA tracks the continuous methods closely.
We note that the differences between the parametric methods are moderate rather than spanning orders of magnitude, and the qualitative distinction lies in the shape of the profiles: smooth decay for the continuous methods against irregular, non-monotone contributions for the grid-based and smoothing-based ones.


\subsection{Computational Cost and Parameter-Count Definitions}
\label{appendix:runtime_eval}

\paragraph{Hardware:}
CPU-based baselines (OMP, SSA, Prophet, MSTL, MCA, Analysis K-SVD, NOMP, SFW) were run on Intel Xeon Gold and AMD EPYC nodes, while gradient-based methods (CDP, N-BEATS) were accelerated on NVIDIA RTX GPUs.

\paragraph{Parameter counting:}
Comparing parameter efficiency across disparate model classes requires an explicit definition of what is counted for each method.
\autoref{tab:param_counting} defines the parameter count used in all Pareto analyses: for every method, we count exactly the set of scalars that must be stored to reproduce its decomposition or reconstruction.

\begin{table}[htbp]
\centering
\caption{\textbf{Parameter-count definitions} used in all Pareto analyses. Each method's count comprises exactly the scalars required to reproduce its output.}
\label{tab:param_counting}
\begin{tabular}{lp{0.62\columnwidth}}
\toprule
\textbf{Method} & \textbf{Counted parameters} \\ \midrule
CDP / CDP (Mod.) & Every continuously optimized scalar: $3$--$5$ atom-specific parameters plus one bias per atom \\
OMP & Frequency index and linear coefficient per selected dictionary atom \\
NOMP & Refined continuous frequency plus complex amplitude per sinusoid \\
SFW & Continuous atom position plus weight per atom \\
MCA & Selected linear coefficients over the grid-discretized sub-dictionaries \\
Analysis K-SVD & All entries of the learned analysis operator rows \\
MSTL & All stored component values (trend and per-period seasonal series) needed to reproduce the decomposition \\
SSA & All entries of the retained singular triples $(\sigma_i, u_i, v_i)$ \\
Prophet & All fitted coefficients (changepoints, trend, and Fourier seasonality terms) \\
N-BEATS & All trainable network weights \\ \bottomrule
\end{tabular}
\end{table}

\begin{figure}[htbp]
  \begin{center}
    \includegraphics[width=\columnwidth]{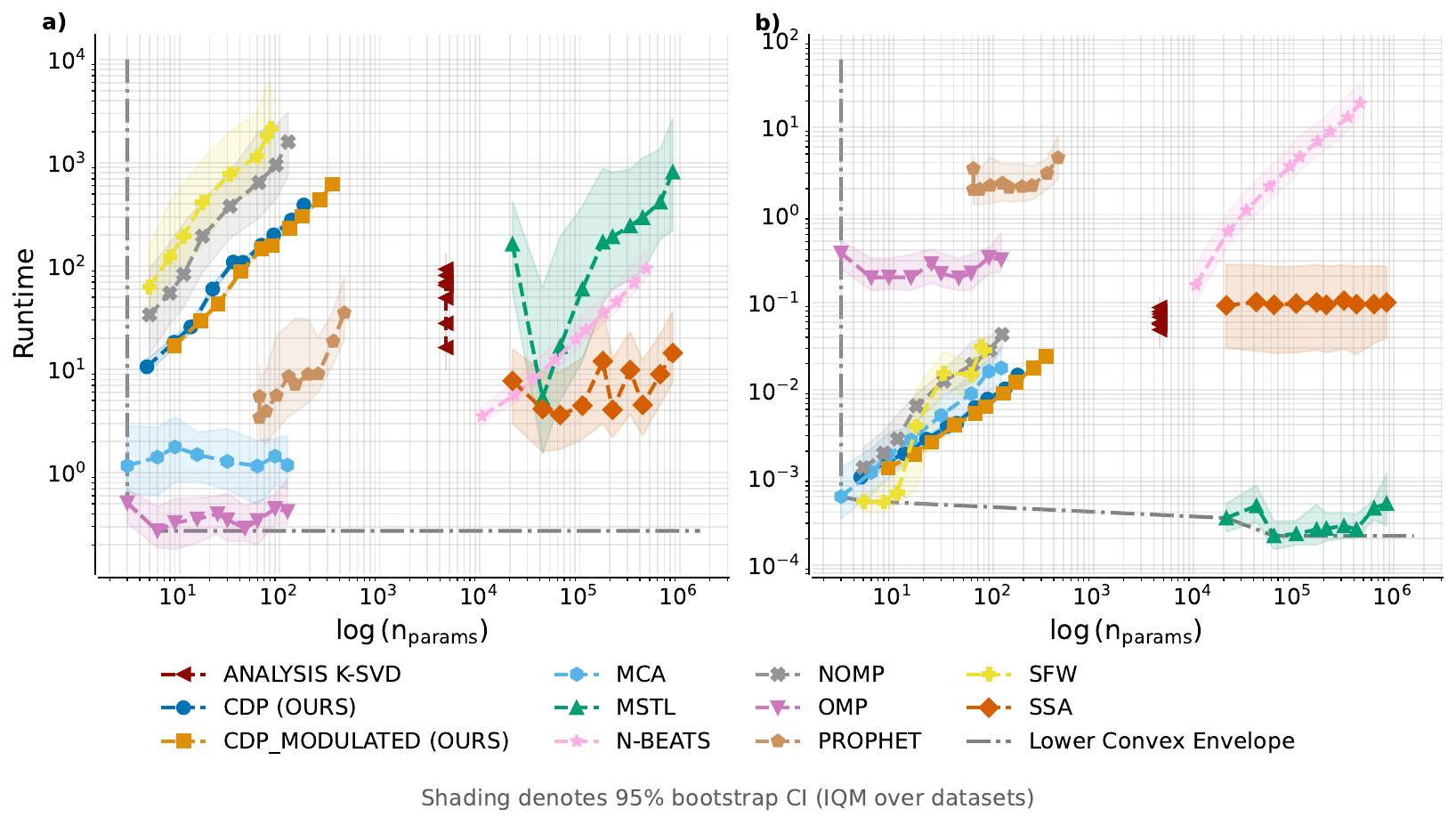}
    \caption{\textbf{Training and inference cost versus model complexity.} \textit{(a)} Fitting time in seconds (log--log scale). \textit{(b)} Evaluation (inference) time in seconds. Shading denotes 95\% bootstrap confidence intervals over datasets.}
    \label{fig:runtime}
  \end{center}
\end{figure}%

\paragraph{Training cost:}
Panel (a) of \autoref{fig:runtime} reports CDP fitting times of approximately $10$s at a single atom rising to $500$s at $40$ atoms. 
At matched atom counts this places CDP one to two orders of magnitude above OMP, a gap that reflects the structural difference in per-atom work. 
OMP evaluates a single inner product against a fixed dictionary at each step, while CDP runs L-BFGS over a continuous parameter manifold initialized from a scout batch of candidates. 
The additional cost is what enables CDP to resolve grid mismatch and recover off-grid parameters, a capability whose impact is documented in \autoref{appendix:synthetic_eval} on signals where OMP fails by construction.
CDP remains substantially below high-parameter baselines such as MSTL, which approaches $10^3$s at $10^6$ parameters.
Within the class of continuous pursuit algorithms, CDP is the fastest at matched atom budgets (\autoref{tab:wallclock}): at $K=40$ its median fitting time is $417$s, against $1722$s for NOMP and $1879$s for SFW, whose standard algorithms re-optimize all atoms in every iteration.
CDP's fitting cost is therefore high relative to classical grid methods but competitive within the class of continuous methods, and it is a one-time cost that yields a closed-form model.
MCA is cheap because it optimizes only linear coefficients on fixed grids, and the cost of the classical and deep baselines is shown in \autoref{fig:runtime}.

\begin{table}[htbp]
\centering
\caption{\textbf{Median fitting time in seconds} per atom budget $K$ for CDP and the continuous and structured baselines, taken over 13 datasets and 10 seeds.}
\label{tab:wallclock}
\begin{tabular}{lrrrrrrrr}
\toprule
\textbf{Method} & $K{=}1$ & $K{=}2$ & $K{=}3$ & $K{=}5$ & $K{=}10$ & $K{=}20$ & $K{=}30$ & $K{=}40$ \\ \midrule
CDP & 12 & 20 & 29 & 63 & 129 & 223 & 329 & 417 \\
NOMP & 22 & 44 & 66 & 125 & 441 & 462 & 695 & 1722 \\
SFW & 46 & 91 & 141 & 216 & 451 & 858 & 1174 & 1879 \\
MCA & 1 & 1 & 1 & 2 & 1 & 1 & 1 & 1 \\
Analysis K-SVD & 95 & 86 & 60 & 73 & 76 & 46 & 30 & 14 \\ \bottomrule
\end{tabular}
\end{table}

\paragraph{Inference cost:}
Panel (b) shows that CDP, once fitted, consists of a small set of closed-form equations and requires only a single forward pass to evaluate, resulting in inference times of $10^{-3}$ to $10^{-2}$s across all atom counts.
This is the lowest inference time of any evaluated method.
OMP requires approximately $0.3$s per query because it must reproject onto its dictionary for longer prediction horizons.
Prophet requires $1$–$3$s per evaluation.
N-BEATS scales from $10^{-1}$s to $10^{1}$s as parameter count grows, making it by far the most expensive at inference despite offering no interpretability advantage.
 
In practice, a decomposition model is fitted once and then used for many downstream queries.
Under this regime, CDP's training cost is a one-time investment, and its near-zero inference latency makes it particularly suitable for real-time or repeated-query applications.


\subsection{Baselines}
\label{appendix:baselines}

This section describes the implementation of all baselines, grouped into classical and neural methods on the one hand and continuous-pursuit and structured-dictionary methods on the other, and reports the analyses that separate dictionary coverage from optimization quality.
For every method, the atom budget $K$ controls the number of components in the decomposition as described below, and the counted parameters are defined in \autoref{tab:param_counting}.

\paragraph{MSTL~\cite{bandara2025_mstl}:}
MSTL decomposes the signal by iterative LoESS smoothing into one trend component and $K-1$ seasonal components, whose periods are set to the $K-1$ dominant frequencies detected via the FFT of the input signal.

\paragraph{SSA~\cite{vautard1992_ssa}:}
SSA embeds the signal into a trajectory matrix and retains the $K$ leading singular triples.
Each component is obtained by Hankel-averaging one rank-one term, and forecasts follow the linear recurrence induced by the retained subspace.

\paragraph{OMP~\cite{pati1993_omp}:}
OMP greedily selects $K$ atoms from a discrete Fourier dictionary and solves for their coefficients by orthogonal projection at each step.

\paragraph{Prophet~\cite{taylor2018_prophet}:}
Prophet fits a piecewise linear trend and $K-1$ Fourier seasonalities, whose periods are set to the $K-1$ dominant FFT-detected frequencies of the input signal.

\paragraph{N-BEATS~\cite{oreshkin2019_nbeats}:}
We use the interpretable N-BEATS configuration with one trend block (polynomial basis) and $K-1$ seasonality blocks (Fourier basis), trained for point-by-point forecasting on a sliding input window.

\paragraph{Newtonized OMP (NOMP)~\cite{mamandipoor2016_nomp}:}
NOMP is a continuous matching pursuit for sinusoids.
Each greedy step selects the strongest frequency on an oversampled FFT grid and refines it over the continuum via Newton steps.
All previously selected frequencies are then refined cyclically, and all amplitudes are updated by an exact least-squares solve.

\paragraph{Sliding Frank-Wolfe (SFW)~\cite{denoyelle2019_sfw}:}
SFW is the standard algorithm for the continuous sparse approximation problem (BLASSO).
Each iteration inserts the atom that maximizes correlation with the residual and then jointly moves all atom positions and weights by local optimization.
Like NOMP, it operates on a single chosen analytic family.

\paragraph{Morphological Component Analysis (MCA)~\cite{starck2005_mca,bobin2007_mca}:}
We instantiate MCA with the same primitive families as CDP (polynomial trend, sinusoids at FFT-detected frequencies, Gaussian transients), but with all atom parameters frozen on discrete grids.
Only the linear coefficients are optimized, by block-coordinate thresholding over the sub-dictionaries.
MCA is therefore a \emph{discretized ablation} of CDP, and any performance gap at matched atom families is attributable to continuous parameter optimization.

\paragraph{Analysis K-SVD~\cite{rubinstein2013_analysisksvd}:}
Analysis K-SVD performs analysis-model dictionary learning with backward-greedy pursuit and eigenvector-based row updates on overlapping patches.
Forecasts are produced by extrapolating the learned annihilation recurrence, protected by divergence guards.
Since the co-sparse model stores the full analysis operator rather than an additive atom set, its parameter count is independent of the atom budget, which only sets the dimension of the per-patch signal subspace.
Analysis K-SVD therefore appears at a single parameter count in the Pareto plots.

\paragraph{Results: two regimes.}
In the sparse regime ($K = 2$ to $5$), CDP attains the best median test error of all methods across the 13 datasets, with $0.65$ against $0.97$ for NOMP and $0.87$ for SFW at $K{=}2$, $0.62$ against $1.05$ and $0.84$ at $K{=}3$, and $0.64$ against $0.93$ for both at $K{=}5$.
At $K = 10$ the continuous methods are close, with $0.76$ for CDP against $0.73$ for NOMP and SFW.
At large budgets ($K \geq 20$), the single-family continuous methods profit from exact update steps available for pure sinusoid dictionaries: NOMP reaches lower training error ($0.23$ against $0.25$ at $K{=}20$) and SFW lower test error ($0.68$ against $0.89$ at $K{=}40$).

\paragraph{Results: matched-family MCA ablation.}
Across all 13 datasets, CDP dominates MCA at every atom budget on training reconstruction (median $0.69$ against $0.73$ at $K{=}1$, $0.43$ against $0.59$ at $K{=}5$, and $0.16$ against $0.34$ at $K{=}40$) and on test error ($0.62$--$0.94$ against $1.3$--$1.7$ across budgets).
Since the atom families are matched by construction, this gap isolates the contribution of continuous parameter optimization.

\paragraph{Results: Analysis K-SVD.}
As expected for a co-sparse patch model, Analysis K-SVD excels at in-sample denoising and reaches a near-zero training residual.
Its learned operator rows are however numerical vectors without closed form, and extrapolation through the learned recurrence is unstable, with a median test error above $28$ across datasets even with divergence guards.
It illustrates what is gained (flexibility) and lost (interpretability, forecastability) when parametric structure is abandoned.

\paragraph{Morphology breakdown: coverage versus optimization quality.}
To separate library coverage from optimizer quality we use CDP restricted to a sinusoid-only library (CDP-sine), which shares its optimizer with full CDP and its dictionary family with NOMP and SFW.
We compare training reconstruction loss at a budget of $5$ atoms (\autoref{tab:morphology_breakdown}).
First, \emph{optimization quality}: on the strongly periodic datasets (PEMS03, PEMS04, PEMS08, Solar, Traffic), CDP-sine matches the best single-family baseline to within $1.2\%$ on four of the five, with $3.7\%$ on Solar as the worst case ($0.167$ against $0.161$ for NOMP).
At a matched dictionary the continuous optimizers are therefore equivalent.
Second, \emph{coverage}: the full library improves reconstruction over CDP-sine on $9$ of $13$ datasets, by $13$--$28\%$ on datasets with switching or spiky structure (Electricity $0.593 \to 0.514$, PEMS07 $0.436 \to 0.313$, Weather $0.599 \to 0.496$).
With this, full CDP attains the best reconstruction on five datasets (Electricity, ETTm2, PEMS04, PEMS07, Weather), excluding the in-sample-only Analysis K-SVD, and lies within $2.2\%$ of NOMP on ETTh1 and ETTm1.
CDP's advantage is thus a coverage effect on mixed-morphology data on top of an optimizer that matches the single-family methods on their home ground.


\subsection{CDP Hyperparameters}
\label{appendix:hyperparams}
 
\autoref{tab:cdp_hyperparams} lists all hyperparameters used in the CDP experiments. The two configurations correspond to CDP and CDP (Modulated), which differ only in the set of atom types available during greedy selection.
 
\begin{table}[htbp]
\centering
\caption{\textbf{CDP hyperparameters} used in all experiments. CDP uses all atom types excluding the ModulatedAtom, while CDP (Modulated) includes it.}
\label{tab:cdp_hyperparams}
\begin{tabular}{p{0.26\columnwidth}p{0.27\columnwidth}p{0.37\columnwidth}}
\toprule
\textbf{Hyperparameter} & \textbf{Value} & \textbf{Description} \\ \midrule
\texttt{scout\_batch\_size} & 10 & Number of candidate atoms evaluated in parallel during truncated optimization \\
\texttt{scout\_epochs} & 5 & Number of L-BFGS steps per candidate during the truncated scouting phase \\
\texttt{global\_epochs} & 100 & Number of L-BFGS steps during full optimization of the selected atom \\
\texttt{num\_global\_optims} & 5 & Number of joint global refinement passes over all active atoms \\
\texttt{n\_atoms} & \{1,2,3,5,8,10,15,20,30,40\} & Atom budgets evaluated for the Pareto frontier \\
\texttt{atom\_types} (CDP) & All except ModulatedAtom & Atom families available for greedy selection \\
\texttt{atom\_types} (CDP Mod.) & All & Atom families available for greedy selection \\
\texttt{lbfgs\_lr} & 1.0 & Learning rate for the L-BFGS optimizer \\
\texttt{lbfgs\_max\_iter} & 20 & Maximum number of iterations per L-BFGS call \\
\texttt{lbfgs\_line\_search\_fn} & Strong Wolfe & Line search method used by L-BFGS \\
\texttt{limit\_loss\_weight} & $10^3$ & Weight applied to soft barrier penalty terms \\ 
\texttt{start\_sharpness} & $1.0$ & Starting $\kappa$ for sharpness annealing \\ 
\texttt{end\_sharpness} & $50.0$ & End $\kappa$ for sharpness annealing \\ 
\texttt{sharpness\_growth\_rate} & $\log(\frac{\kappa_{\text{end}}}{\kappa_\text{start}})$ & Growth rate $\gamma$ for sharpness annealing \\ 
\bottomrule
\end{tabular}
\end{table}


\subsection{Individual Evaluations per Dataset}
\label{appendix:individual_dataset}

The following figures report reconstruction and forecasting MSE as a function of parameter count for each individual dataset in the LTSF benchmark suite, including all baselines described in \autoref{appendix:baselines}.
These per-dataset results underlie the aggregated Pareto frontiers shown in the main paper.

\begin{figure}[htbp]
  \begin{center}
    \includegraphics[width=\columnwidth, clip]{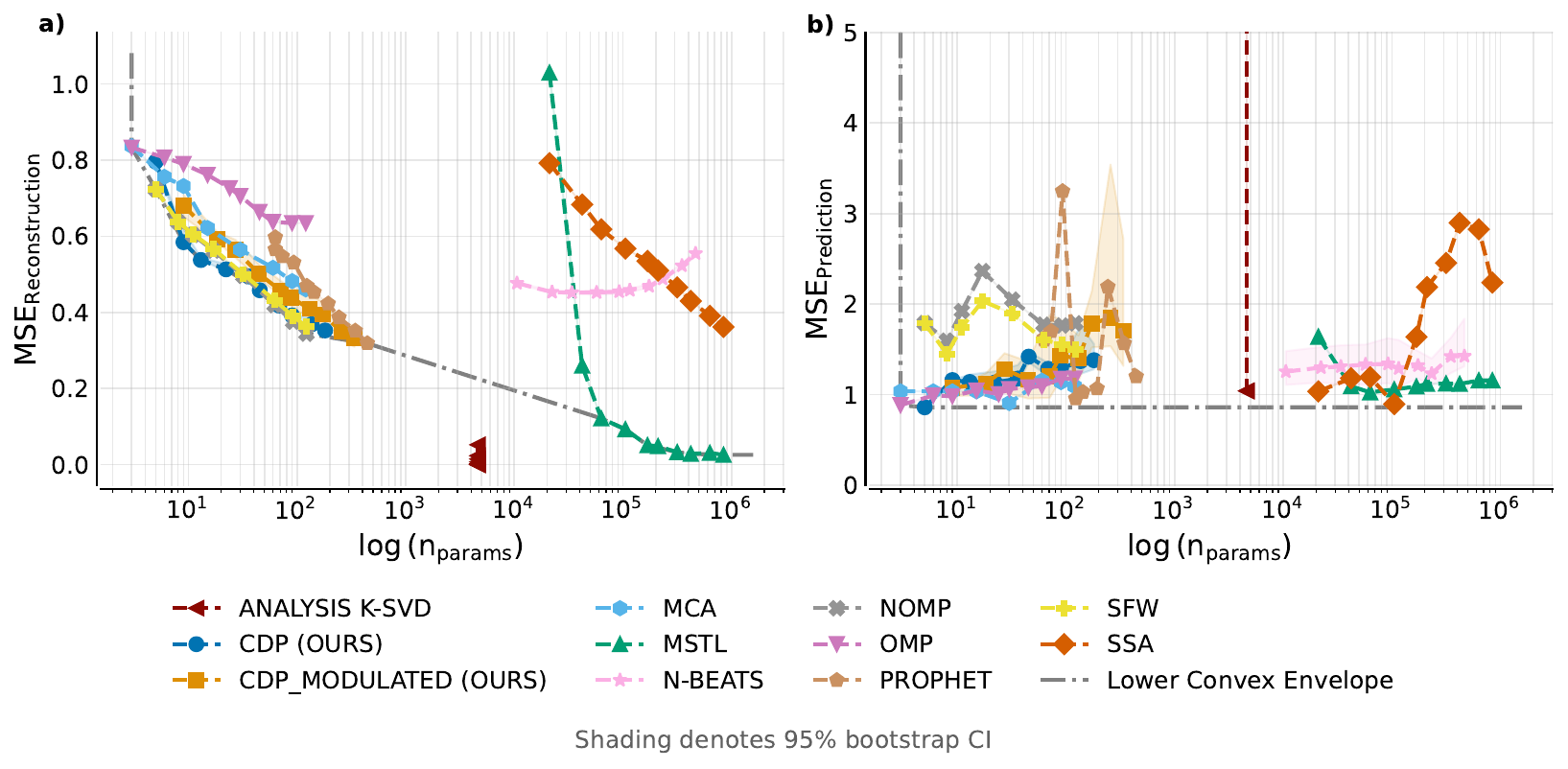}
    \caption{\textbf{Electricity.} Reconstruction (a) and forecasting (b) MSE versus parameter count.}
  \end{center}
\end{figure}%
\begin{figure}[htbp]
  \begin{center}
    \includegraphics[width=\columnwidth, clip]{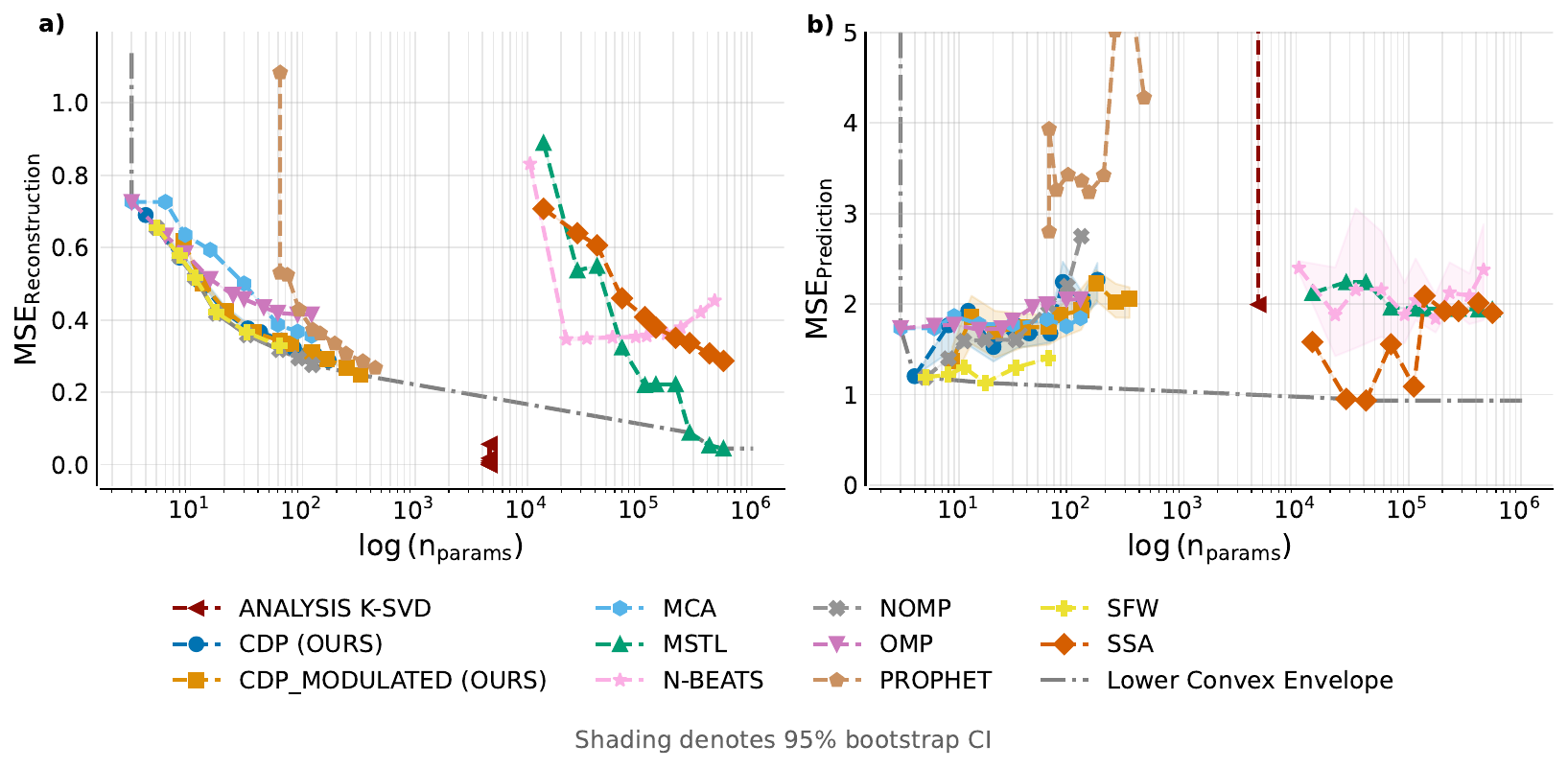}
    \caption{\textbf{ETTh1.} Reconstruction (a) and forecasting (b) MSE versus parameter count.}
  \end{center}
\end{figure}%
\begin{figure}[htbp]
  \begin{center}
    \includegraphics[width=\columnwidth, clip]{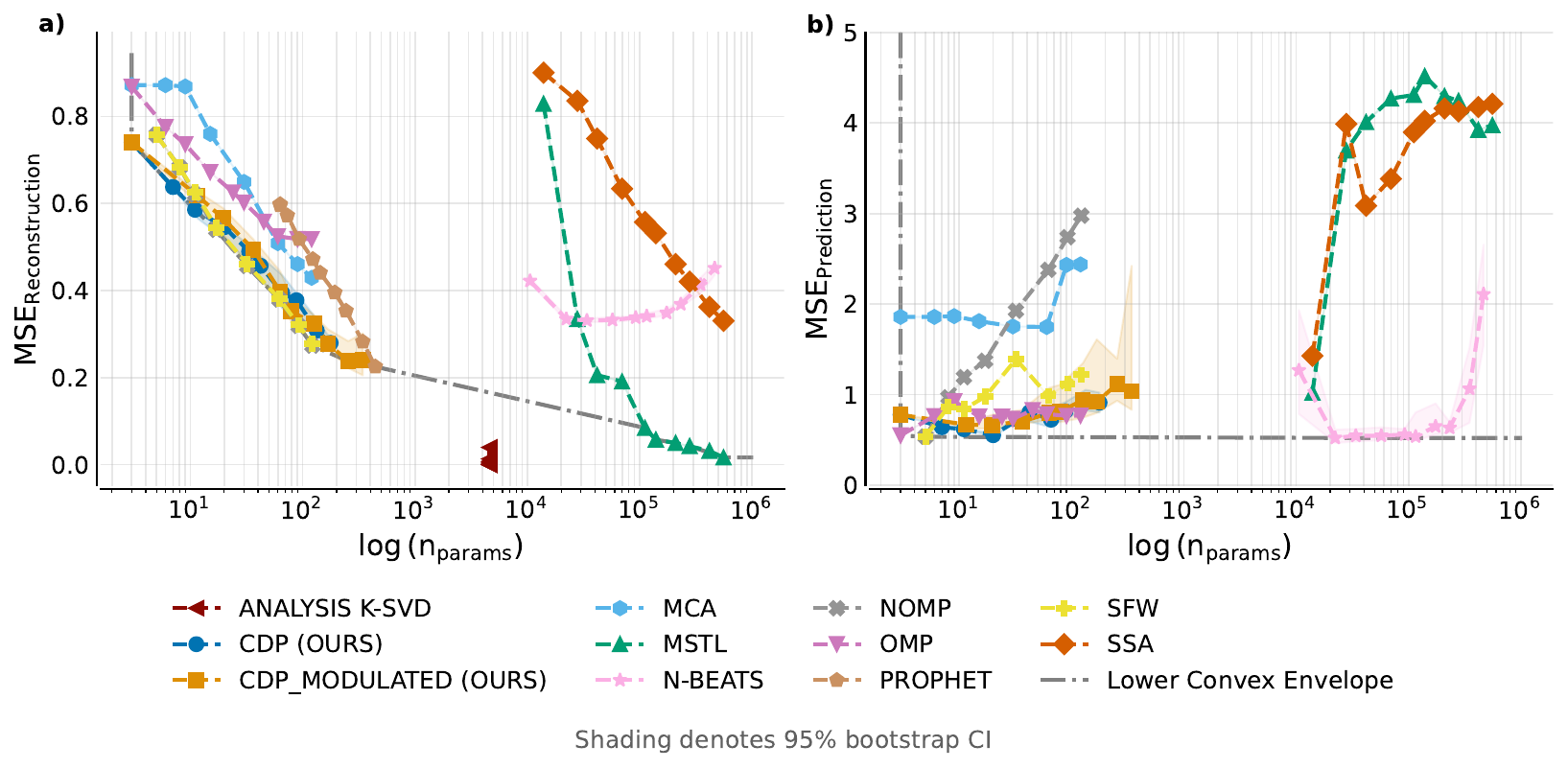}
    \caption{\textbf{ETTh2.} Reconstruction (a) and forecasting (b) MSE versus parameter count.}
  \end{center}
\end{figure}%
\begin{figure}[htbp]
  \begin{center}
    \includegraphics[width=\columnwidth, clip]{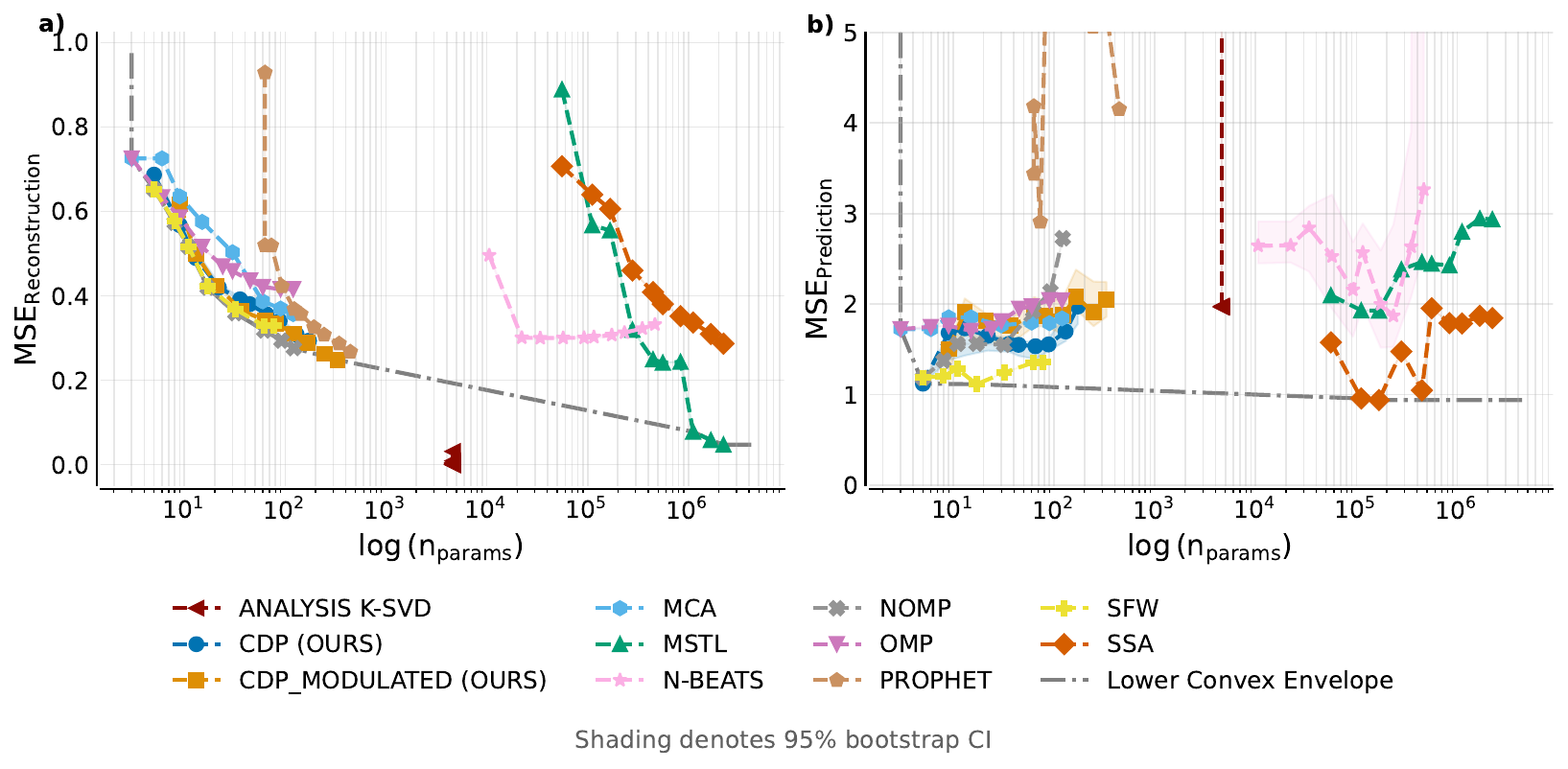}
    \caption{\textbf{ETTm1.} Reconstruction (a) and forecasting (b) MSE versus parameter count.}
  \end{center}
\end{figure}%
\begin{figure}[htbp]
  \begin{center}
    \includegraphics[width=\columnwidth, clip]{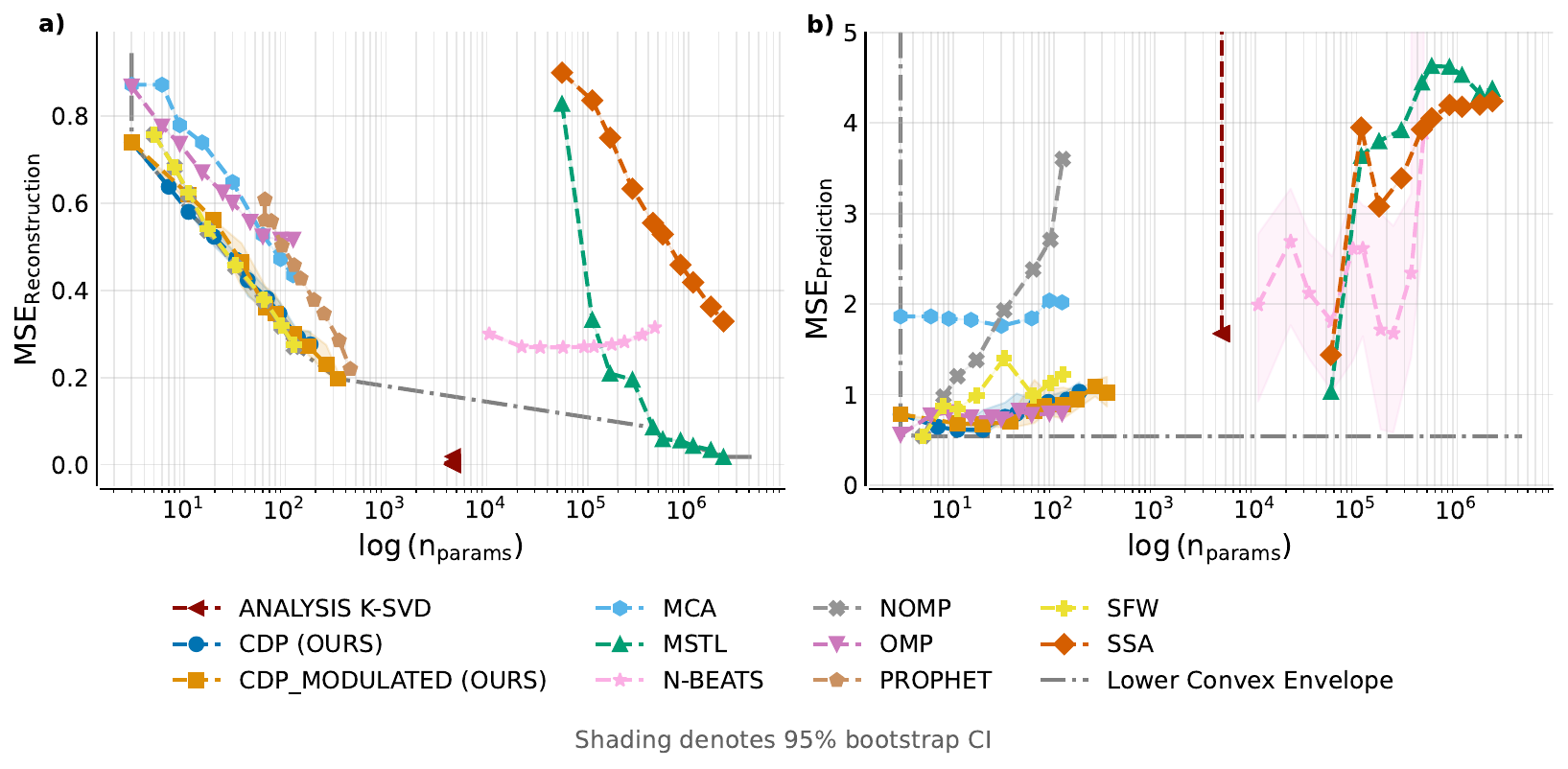}
    \caption{\textbf{ETTm2.} Reconstruction (a) and forecasting (b) MSE versus parameter count.}
  \end{center}
\end{figure}%
\begin{figure}[htbp]
  \begin{center}
    \includegraphics[width=\columnwidth, clip]{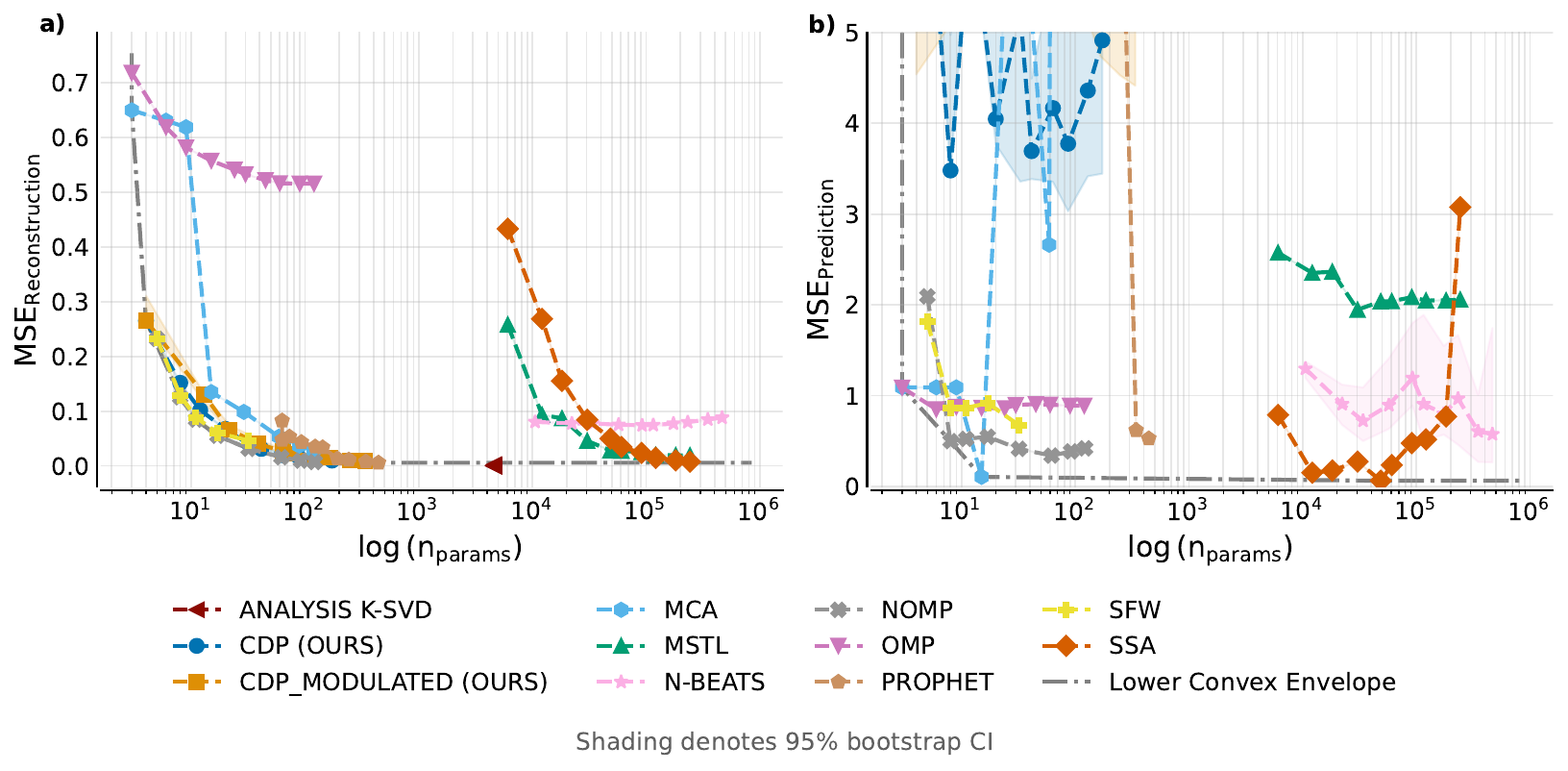}
    \caption{\textbf{Exchange.} Reconstruction (a) and forecasting (b) MSE versus parameter count.}
  \end{center}
\end{figure}%
\begin{figure}[htbp]
  \begin{center}
    \includegraphics[width=\columnwidth, clip]{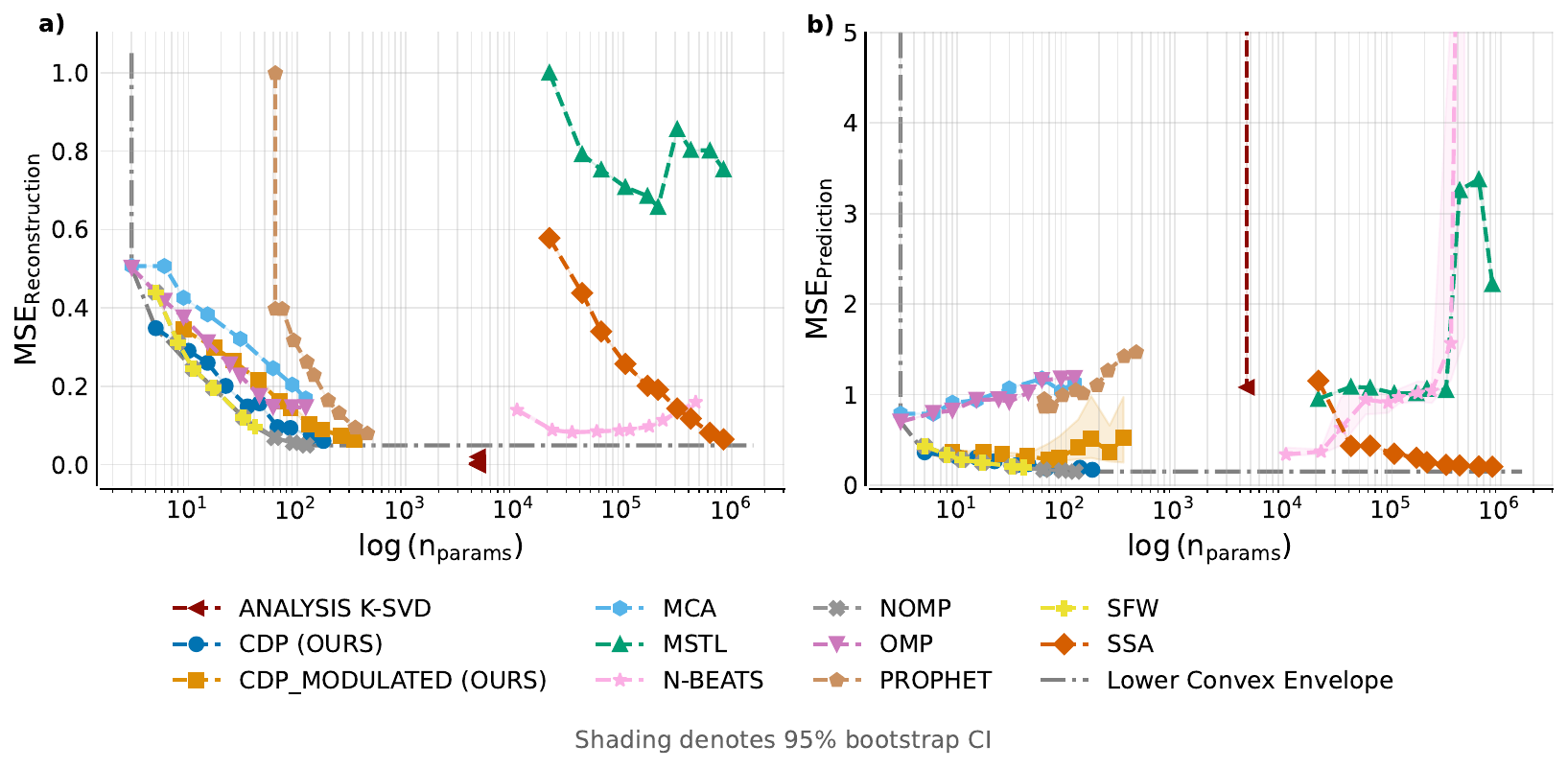}
    \caption{\textbf{PEMS03.} Reconstruction (a) and forecasting (b) MSE versus parameter count.}
  \end{center}
\end{figure}%
\begin{figure}[htbp]
  \begin{center}
    \includegraphics[width=\columnwidth, clip]{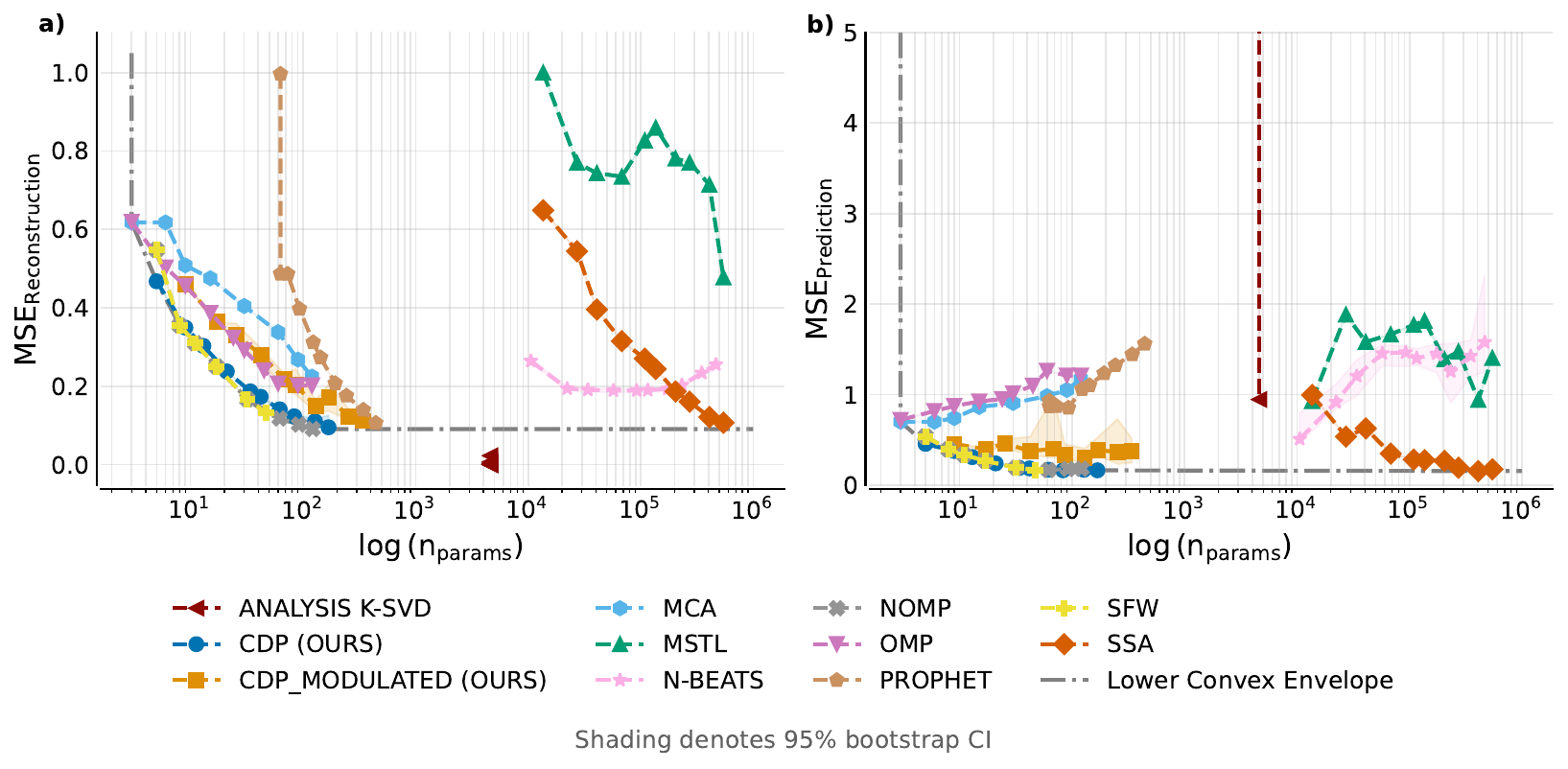}
    \caption{\textbf{PEMS04.} Reconstruction (a) and forecasting (b) MSE versus parameter count.}
  \end{center}
\end{figure}%
\begin{figure}[htbp]
  \begin{center}
    \includegraphics[width=\columnwidth, clip]{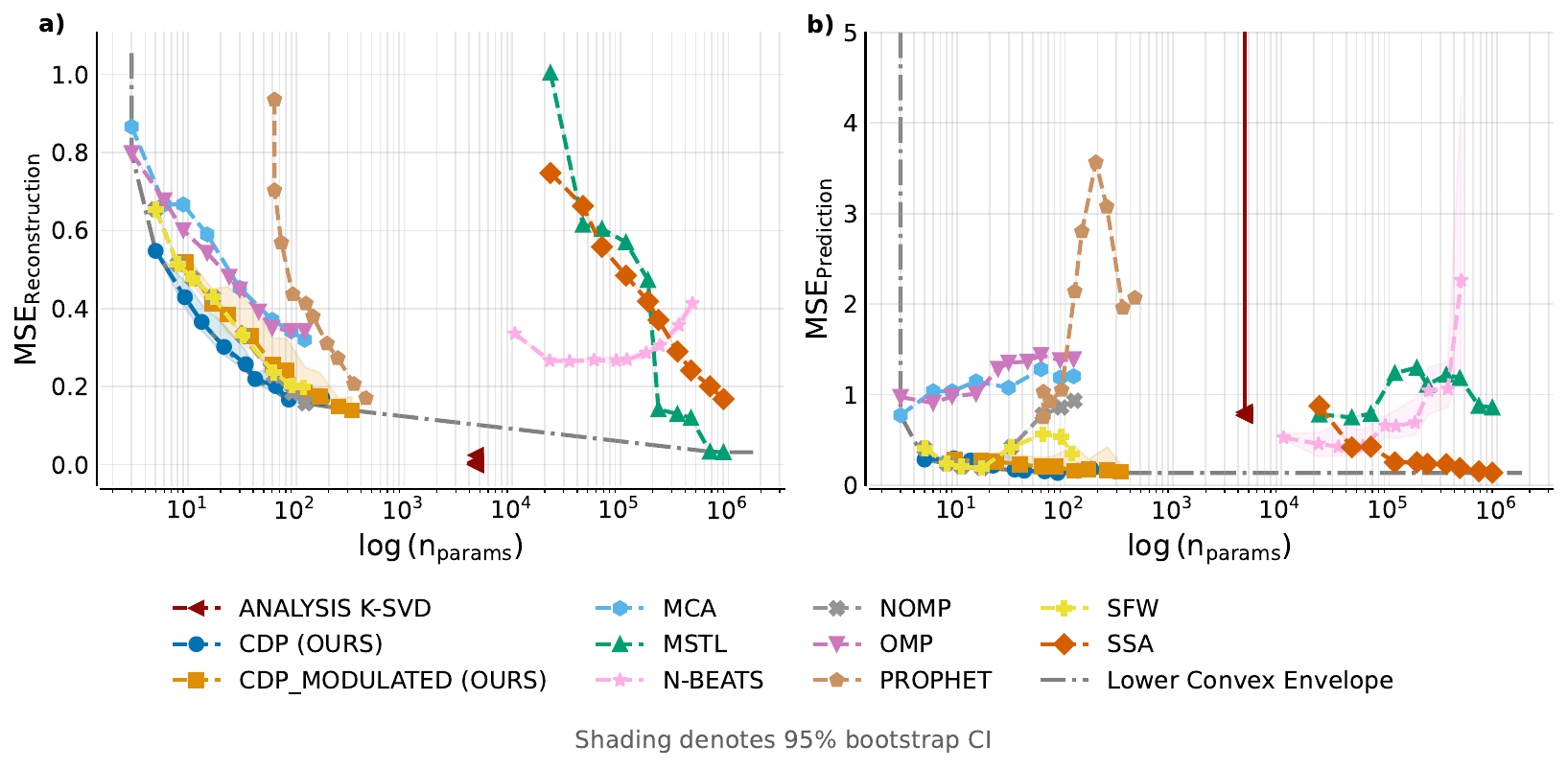}
    \caption{\textbf{PEMS07.} Reconstruction (a) and forecasting (b) MSE versus parameter count.}
  \end{center}
\end{figure}%
\begin{figure}[htbp]
  \begin{center}
    \includegraphics[width=\columnwidth, clip]{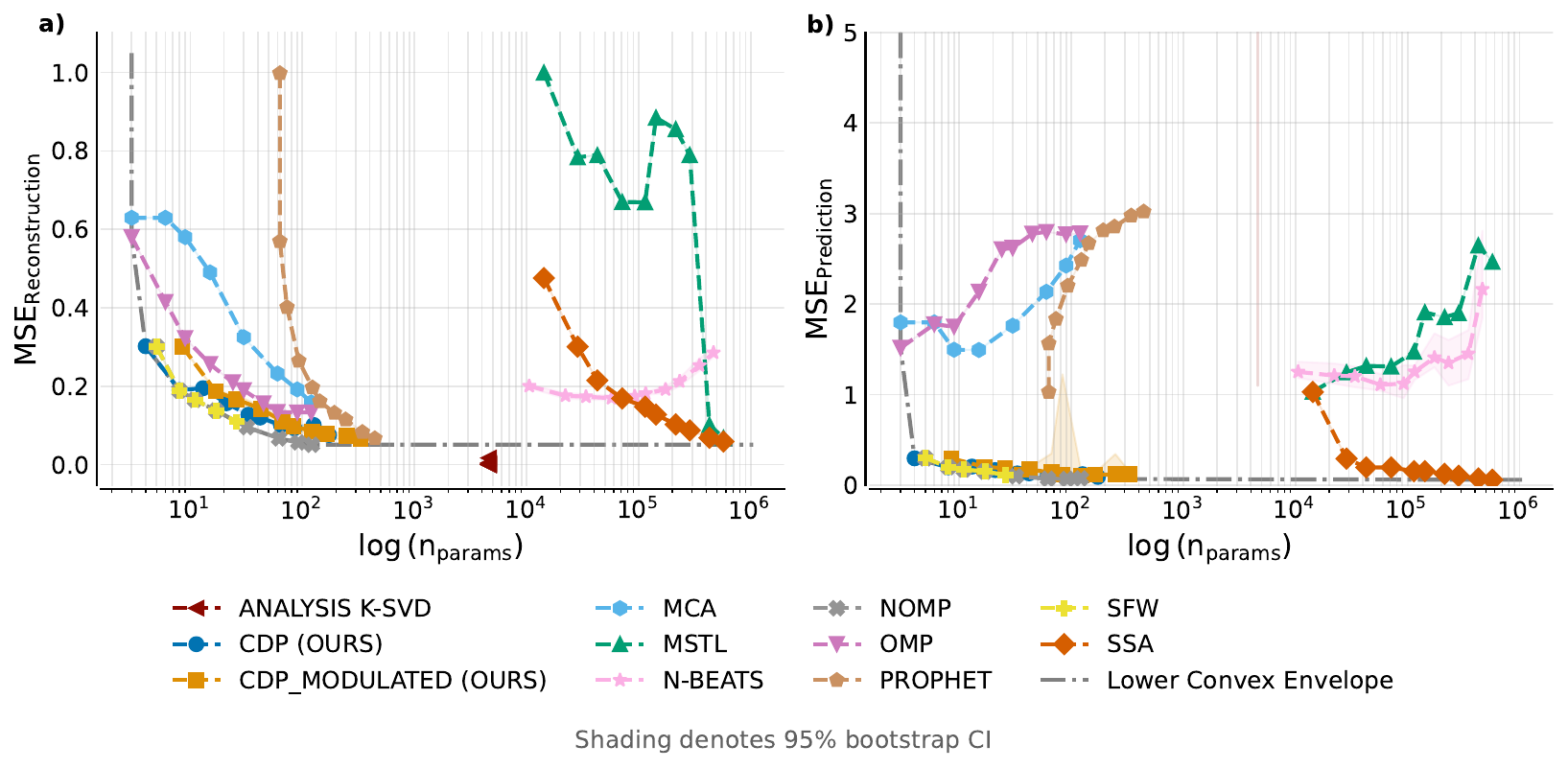}
    \caption{\textbf{PEMS08.} Reconstruction (a) and forecasting (b) MSE versus parameter count.}
  \end{center}
\end{figure}%
\begin{figure}[htbp]
  \begin{center}
    \includegraphics[width=\columnwidth, clip]{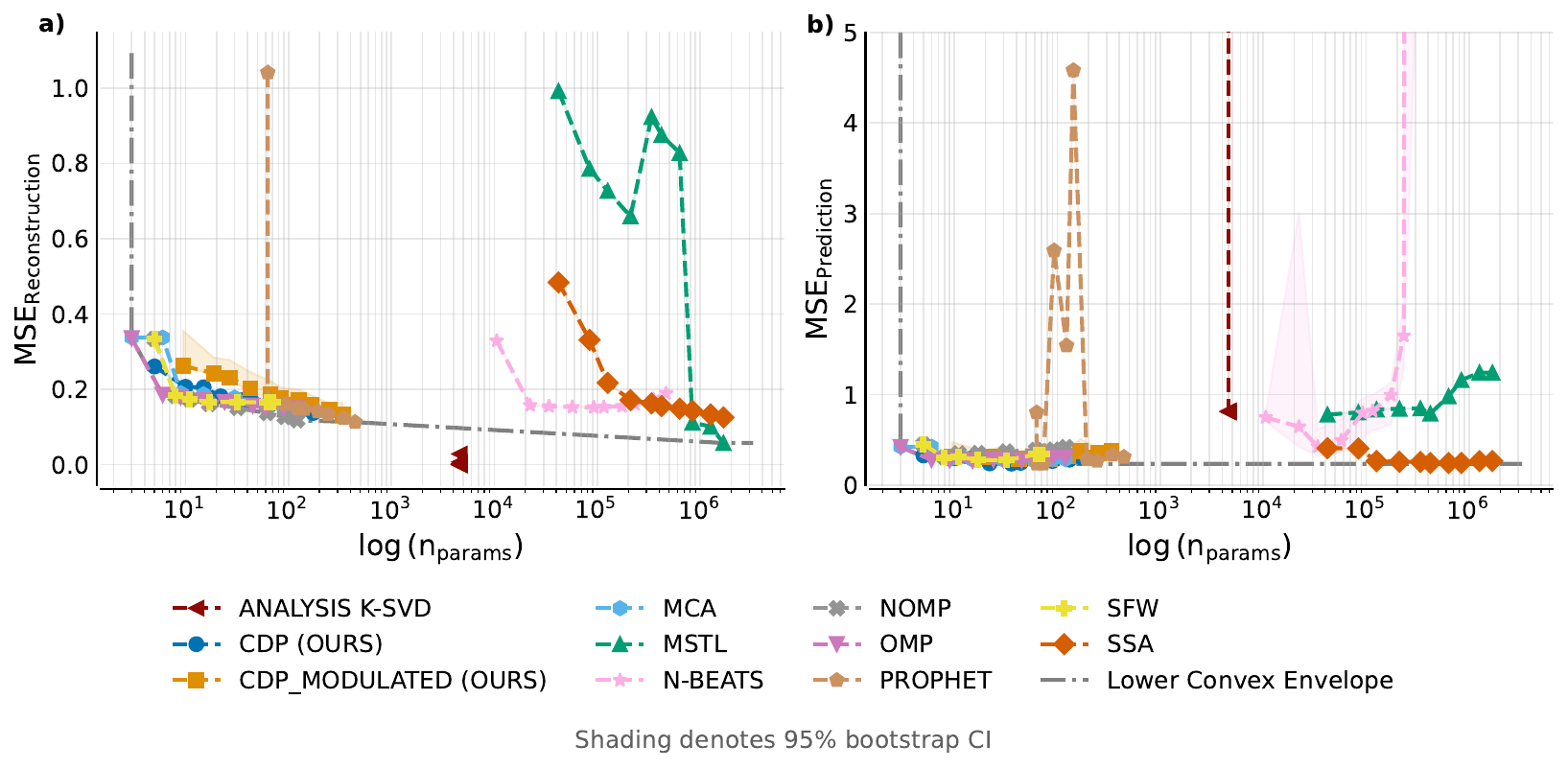}
    \caption{\textbf{Solar.} Reconstruction (a) and forecasting (b) MSE versus parameter count.}
  \end{center}
\end{figure}%
\begin{figure}[htbp]
  \begin{center}
    \includegraphics[width=\columnwidth, clip]{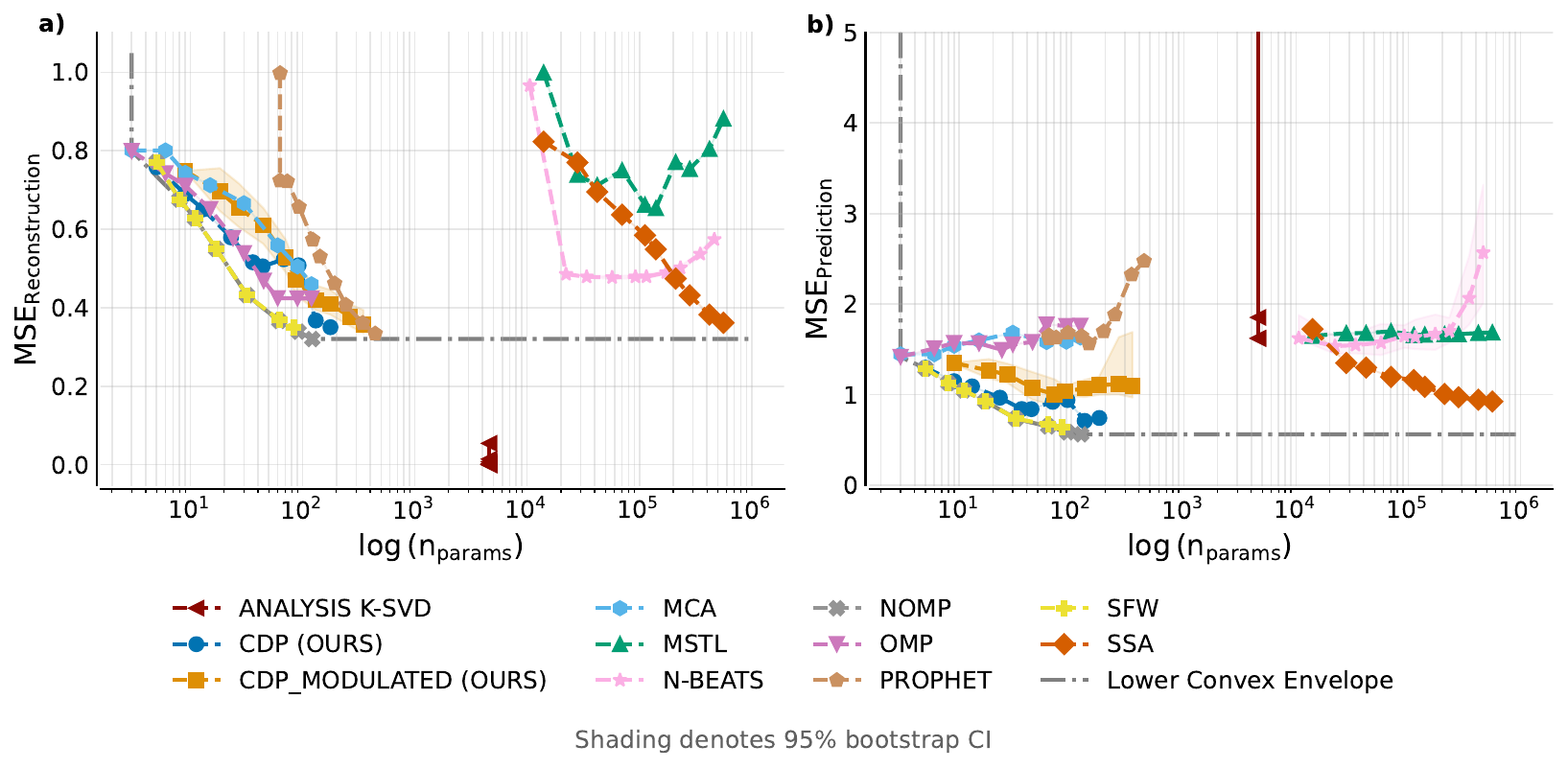}
    \caption{\textbf{Traffic.} Reconstruction (a) and forecasting (b) MSE versus parameter count.}
  \end{center}
\end{figure}%
\begin{figure}[htbp]
  \begin{center}
    \includegraphics[width=\columnwidth, clip]{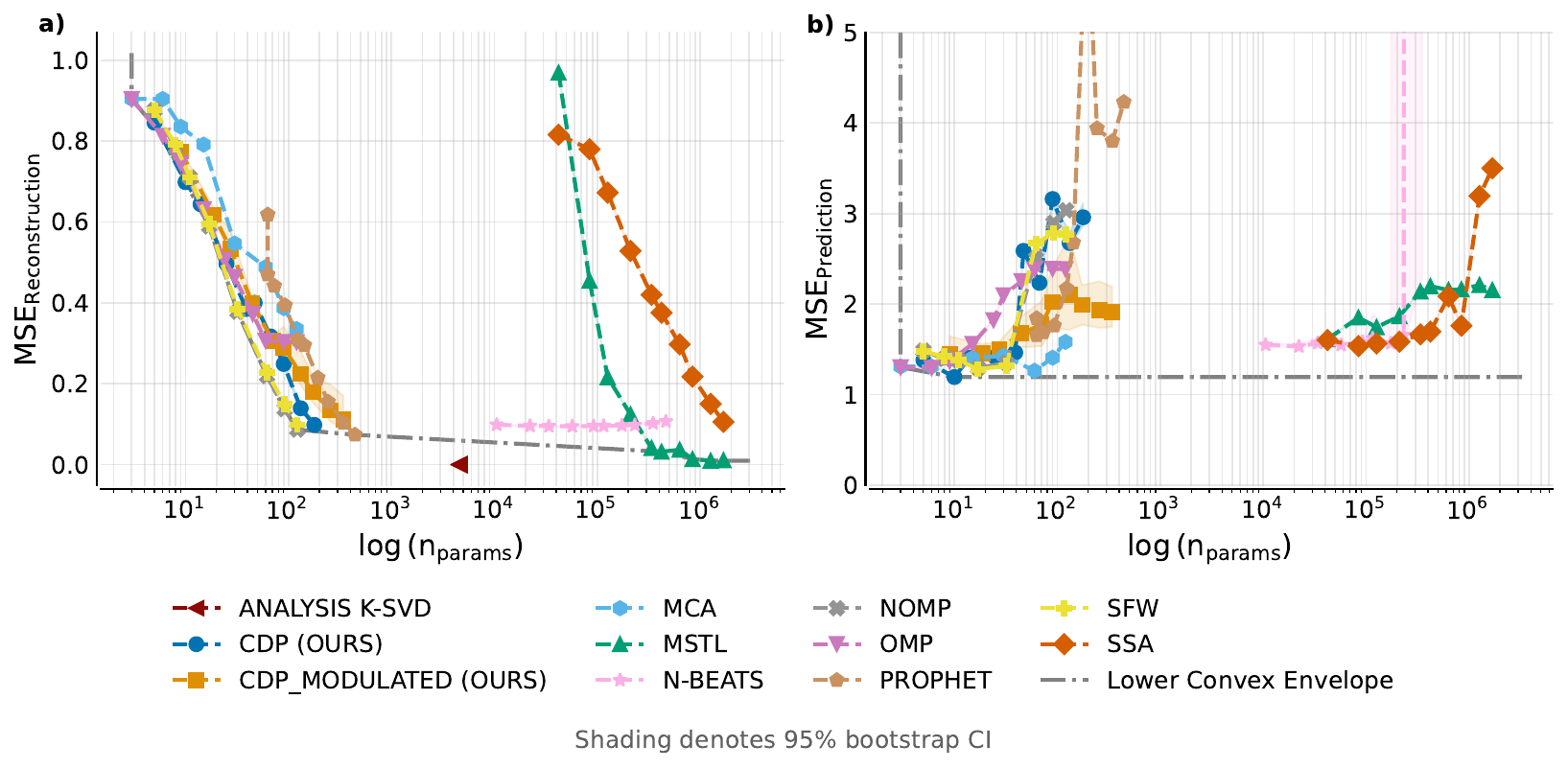}
    \caption{\textbf{Weather.} Reconstruction (a) and forecasting (b) MSE versus parameter count.}
  \end{center}
\end{figure}%



\end{document}